\documentclass[11pt,letterpaper]{article}

\usepackage{amsmath,amssymb,amsthm}
\usepackage{deepthink}
\usepackage{subcaption}
\usepackage{bm}
\usepackage{adjustbox}
\usepackage{booktabs}
\usepackage{multirow}
\usepackage[nameinlink]{cleveref}
\usepackage[
  backend=biber,
  style=alphabetic,
  maxbibnames=100,
  minbibnames=100,
  maxcitenames=2,
  mincitenames=2
]{biblatex}
\title{Coarse-to-Fine Hierarchical Alignment for UAV-based Human Detection using Diffusion Models}

\newcommand{\jointfirst}{\textsuperscript{\dag}}
\newcommand{\corrauth}{\textsuperscript{\ddag}}

\authorblock{
  \href{https://www.linkedin.com/in/wdli/}{\textbf{Wenda Li}}\jointfirst,
  \textbf{Meng Wu}\jointfirst\textsuperscript,
  \textbf{Liangzhao Chen},
  \textbf{Sungmin Eum}\textsuperscript{1},
  \textbf{Heesung Kwon}\textsuperscript{1},
  \href{https://qingqu.engin.umich.edu/}{\textbf{Qing Qu}}\corrauth
}

\affiliation{
  University of Michigan \quad $\cdot$ \quad \textsuperscript{1}DEVCOM Army Research Laboratory
}

\authornote{
  \jointfirst\ Joint first author \quad
  \corrauth\ Corresponding author
}

\abstracttext{
\noindent UAV-based human detection is severely bottlenecked by the scarcity of annotated data, a problem exacerbated by highly variable deployment conditions such as fluctuating altitudes and dynamic viewpoints. While synthetic data provides a scalable alternative, the profound Sim2Real domain gap severely hinders real-world deployment. To overcome this, we introduce Coarse-to-Fine Hierarchical Alignment (CFHA), a diffusion-based framework that translates synthetic images into the target real domain while strictly preserving original geometric labels. CFHA explicitly decouples global style and local content discrepancies, bridging the domain gap through a three-stage pipeline: (1) Global Style Transfer, which aligns color, illumination, and texture using a few-shot real reference set; (2) Local Refinement, which leverages a super-resolution diffusion model to restore photorealistic, fine-grained details to small human instances; and (3) Hallucination Removal, which filters out generated instances that statistically deviate from the target real-world distribution. Empirically, our comprehensive experiments on public UAV Sim2Real benchmarks demonstrate that the proposed CFHA outperforms non-transformed baselines, achieving average $+7.3$ improvements on mAP@50 and $+4.1$ on mAP@95 across four UAV-based human detection datasets. Our ablation studies further confirm that hierarchical, instance-aware alignment is critical for maximizing robust detection.
}


\date{\today}
\correspondence{\href{mailto:wymond@umich.edu}{wymond@umich.edu}}
\resources{\href{https://github.com/liwd190019/cfha_inner}{Project Page} \quad $\cdot$ \quad \href{https://github.com/liwd190019/cfha_inner}{Code}}

\headerlogo{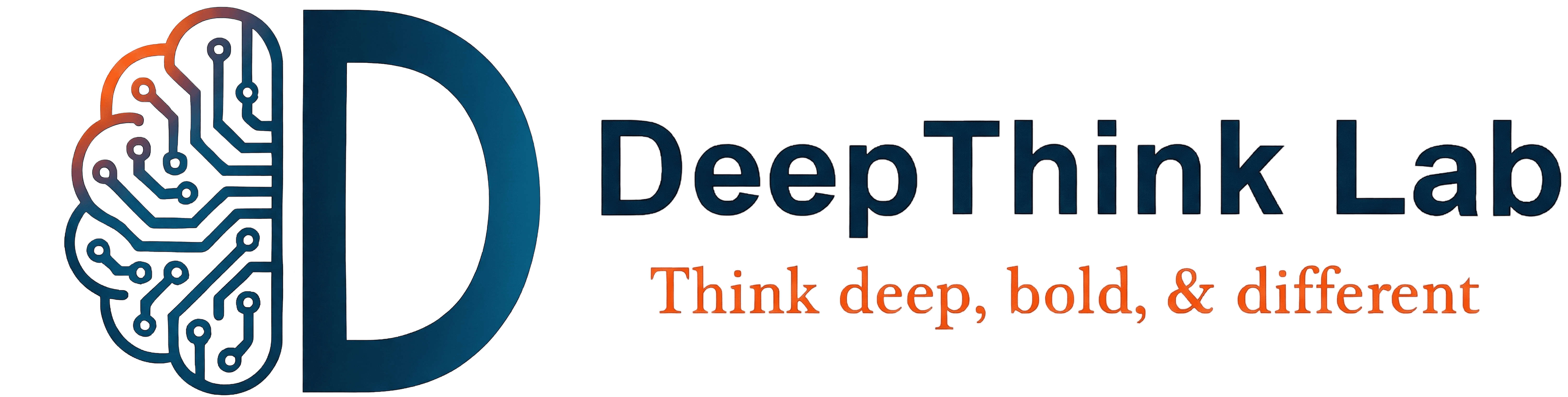}{https://deepthink-umich.github.io}

\begin{document}

\makeDeepthinkHeader


\begin{figure}[h]
  \centering
  \begin{subfigure}[t]{0.38\linewidth}
    \centering
    \includegraphics[width=\linewidth]{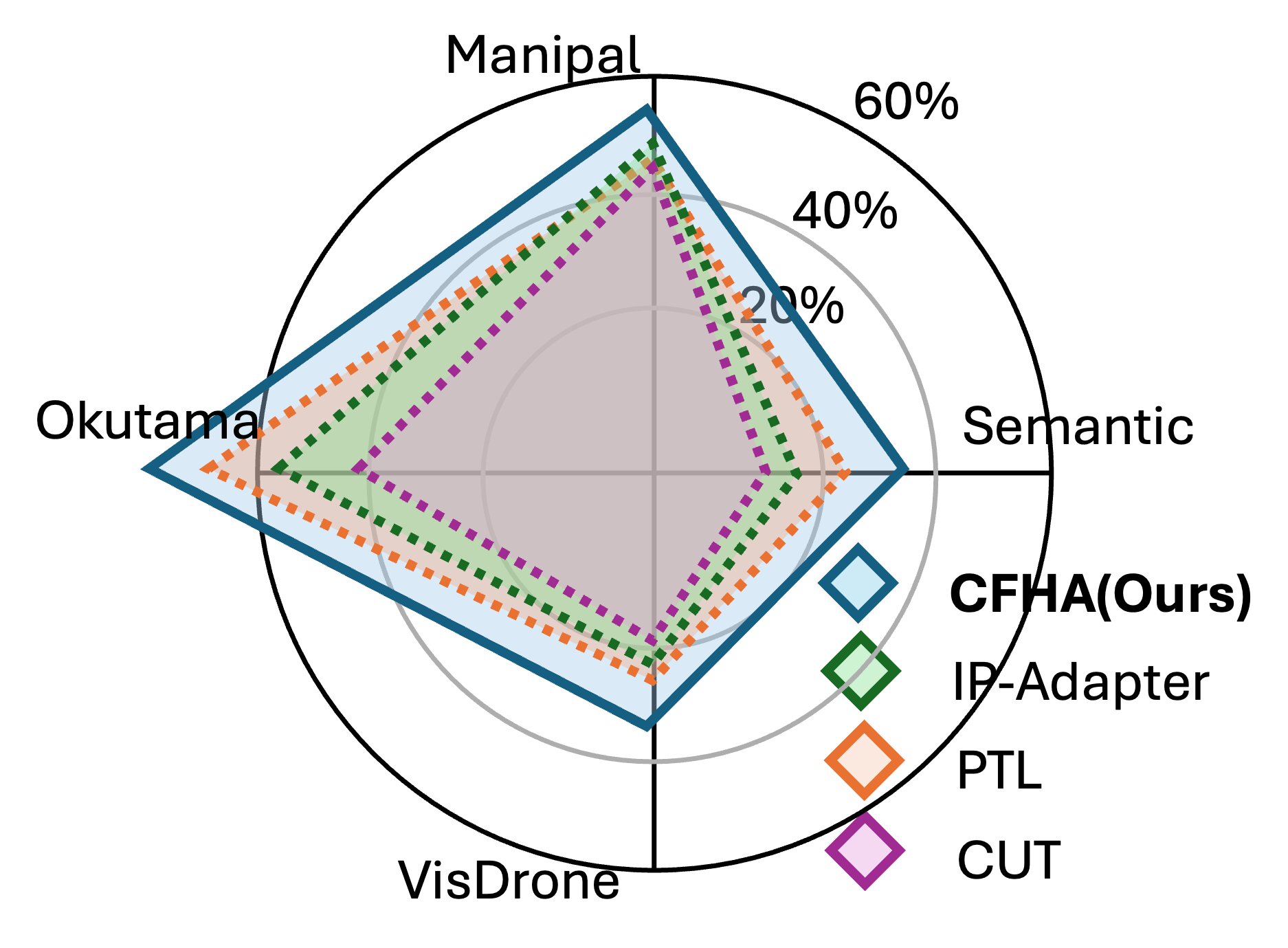}
    \caption{Detector performance (mAP@50) under different data augmentation methods}
    \label{fig:img1}
  \end{subfigure}\hfill
  \begin{subfigure}[t]{0.60\linewidth}
    \centering
    \includegraphics[width=\linewidth]{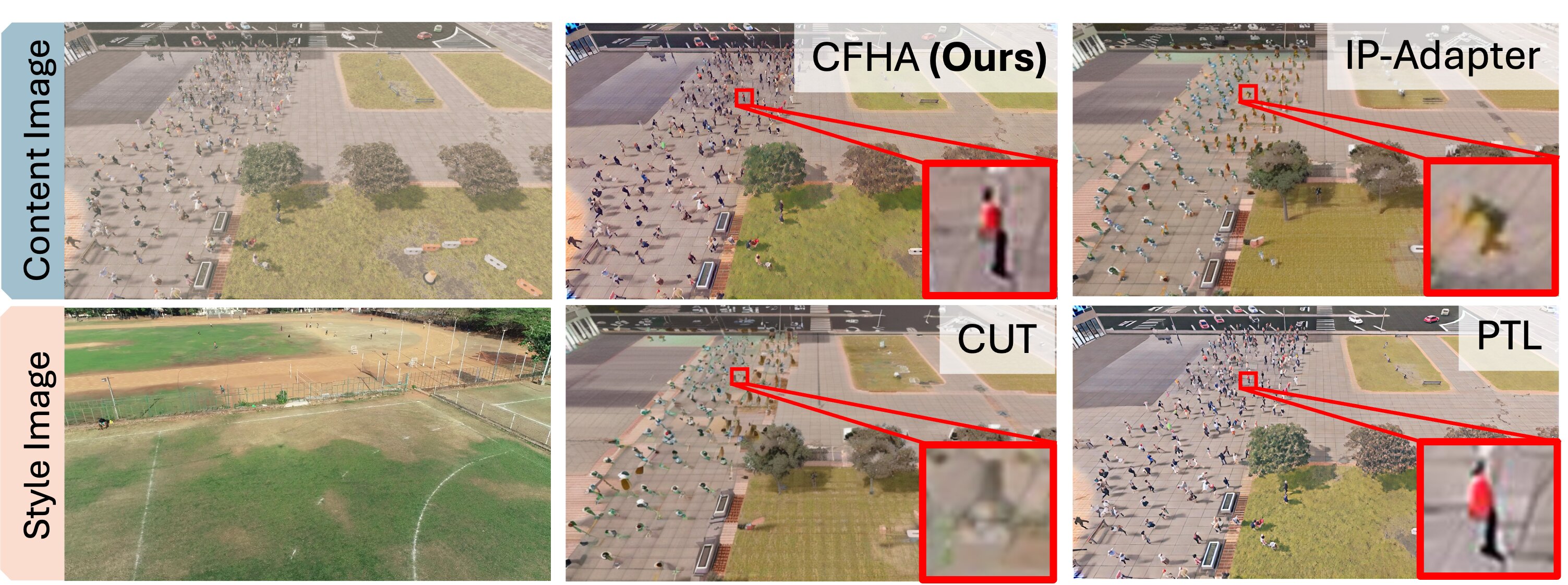}
    \caption{Visual comparison of different data augmentation methods.}
    \label{fig:img2}
  \end{subfigure}
  \caption{Quantitative and qualitative comparison of different data augmentation methods.}
  \label{fig:two_side_by_side}
  \vspace{-0.2in}
\end{figure}

\newpage
\tableofcontents

\newpage
\section{Introduction}
\label{sec:intro}
Human detection is a critical task across numerous computer vision and robotics applications, including search and rescue, disaster response, and urban surveillance.  While deep learning has driven remarkable advances in human object detection on standard benchmarks like COCO~~\autocite{DBLP:conf/eccv/LinMBHPRDZ14} and Open-Images~~\autocite{kuznetsova2020open}, these models rely on prohibitively expensive, large-scale annotations (e.g., over 25,000 worker-hours for COCO). This reliance creates a severe bottleneck for Unmanned Aerial Vehicle (UAV) applications. Human detection in UAV imagery is inherently a long-tailed, low-data vision task; it is characterized by extreme variations in scale, viewpoint, and background, alongside the high cost of collecting and annotating real-world aerial data.  Consequently, this sheer scarcity of accurately labeled training data significantly degrades the effectiveness of standard deep learning-based detectors.

To overcome this data scarcity, training models on synthetic data has emerged as a highly attractive solution. Modern game engines ~\autocite{sanders2016introduction} and generative models ~\autocite{goodfellow2020generative, rombach2022high} can render massive volumes of accurately labeled data at a negligible marginal cost~~\autocite{DBLP:conf/cvpr/TremblayPABJATC18,DBLP:conf/nips/FuTSC0DI23}. However, this potential is severely undermined by the notorious \emph{sim-to-real} (Sim2Real) domain gap. Detectors trained exclusively on synthetic data frequently fail during real-world deployment due to critical discrepancies in lighting, texture, and low-level appearance~~\autocite{DBLP:journals/corr/abs-1806-09755}. Bridging this gap remains a formidable open challenge, particularly given the highly variable conditions unique to UAV imagery.

To address this bottleneck, state-of-the-art diffusion models have emerged as powerful tools for Sim2Real style transfer, leveraging their capacity to generate photorealistic images while preserving underlying content~~\autocite{DBLP:conf/cvpr/RombachBLEO22, DBLP:conf/iclr/PodellELBDMPR24}. However, their deployment in critical domains like UAV-based object detection is impeded by two primary hurdles. First, existing diffusion-based adaptation methods are highly data-dependent, requiring abundant real-world images for fine-tuning. This renders them impractical for the common few-shot scenarios where target-domain data is scarce. Second, standard style transfer techniques frequently fail on UAV imagery due to the prevalence of small objects, such as human pedestrians. During the global translation process, diffusion models often corrupt the semantic structures of these small objects, hallucinating severe artifacts and destroying the precise features required for robust detection.

Fundamentally, these failures stem from a limited understanding of the Sim2Real domain gap. In this work, we identify and formalize two complementary factors of this gap that critically affect detector training: (i) the \emph{global style gap}, including differences in lighting, texture statistics, and camera pose; and (ii) the \emph{local content gap}, capturing discrepancies in object-level characteristics such as instance density and spatial arrangement. Notably, local content inconsistencies are often introduced during the global transformations, resulting in severe hallucinations that introduce spurious object patterns. While existing Sim2Real methods focus almost exclusively on closing the global style gap ~\autocite{zhang2019multimodal, chen2021understanding}, our key insight is that the local content gap is an equally critical, yet entirely overlooked, source of performance degradation. For instance, a synthetic image might contain hundreds of simulated pedestrians, whereas a real-world drone image typically features fewer than ten. This sharp contrast in human instance density represents a fundamental structural mismatch that global style transfer inherently fails to address. For a more detailed discussion of related works, we refer interested readers to .

\begin{figure*}[t]
    \centering
    \begin{adjustbox}{width=\textwidth,center}     \includegraphics{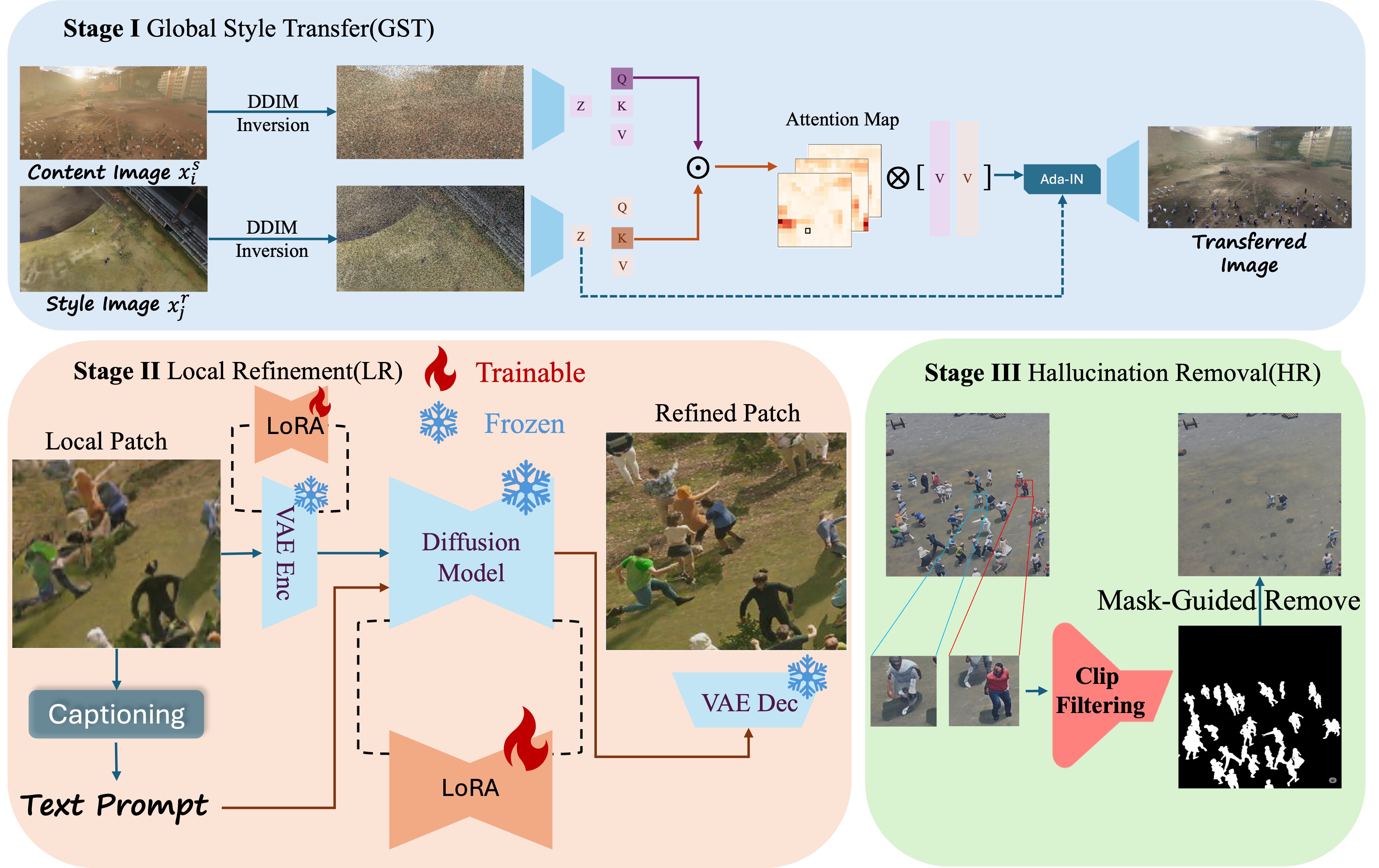}
    \end{adjustbox}
    \vspace{-0.2in}
    \caption{\small \textbf{Overview of our three-stage synthetic-to-real translation pipeline.}  
}
    \label{fig:pipeline}

\vspace{-0.2in}
\end{figure*}

\smallskip 
\noindent \textbf{Our contributions.} To overcome these challenges, we propose Coarse-to-Fine Hierarchical Alignment (CFHA). This novel framework systematically bridges both style and content gaps in a few-shot setting, uniquely preserving the high-fidelity details of small, distant objects. As in \Cref{fig:pipeline},  our method operates in three synergistic stages: (i) \underline{Stage I: Global Style Transfer (GST)} performs style alignment in the latent space via diffusion inversion. A structure-aware cross-attention mechanism computes attention weights to extract and fuse aligned style features with content features, yielding a globally styled image that preserves structural fidelity. (ii) \underline{Stage II: Local Refinement (LR)} enhances local fidelity by cropping object-centric patches and generating corresponding text prompts via image captioning. Conditioned on these prompts, we leverage a pretrained diffusion model—fine-tuned via low-rank adaptation (LoRA) ~\autocite{hu2022lora,yaras2024compressible} to refine the visual quality of each patch. (iii) \underline{Stage III: Hallucination Removal (HR)} addresses artifacts through CLIP-based semantic filtering and mask-guided deletion, ensuring data-consistent final outputs. In summary, our contributions are:
\begin{itemize}[leftmargin=*]
\item \textbf{Unified global--local alignment.} We introduce CFHA, a three-stage framework that addresses the Sim2Real gap from both global and instance-level perspectives. While most existing Sim2Real approaches primarily focus on global appearance transformations.
\vspace{0.05in}
\item \textbf{Few-shot diffusion-based image translation.}
We develop a few-shot Sim2Real adaptation framework based on diffusion-driven image-to-image translation.
\vspace{0.05in}
\item \textbf{State-of-the-art performance.} Extensive evaluations confirm that CFHA- outperform existing Sim2Real baselines. 
\end{itemize}

\section{Preliminary and Related Works}

This section establishes the problem formulation, mathematical notation, and relevant prior work.

\subsection{UAV-based Human Detection with Sim2Real Adaptation}

\noindent \textbf{UAV-based human object detection} aims to locate human objects in imagery captured by UAV. Formally, given an image $\bm{x} \in \mathbb{R}^{H \times W \times 3}$, a human detection model predicts a set of bounding boxes 
\[
\mathcal{B} = \{(\bm{a}_k, \bm{b}_k, \bm{w}_k, \bm{h}_k)\}_{k=1}^{M},
\]
where $(\bm{a}_k, \bm{b}_k)$ represents the center of the $k$-th detected human, and $(\bm{w}_k, \bm{h}_k)$ denotes the width and height of the bounding box. 
Optionally, the detector outputs confidence scores $\{\bm{s}_k\}_{k=1}^{M}$ indicating the likelihood that each bounding box corresponds to a human.

\medskip
\noindent \textbf{Training with synthetic data.}
To address data scarcity in UAV perception in practice, synthetic images are commonly employed to facilitate low-cost model training, where prior works employ advanced algorithms ~\autocite{wang2024convolution, liu2020uav} or large-scale synthetic datasets with free annotations ~\autocite{shen2023progressive, chen2024sim2real}. Formally, let $\mathcal{S}_s = \{(\bm{x}_i^s, \mathcal{B}_i^s)\}_{i=1}^{N_s}$ denote the labeled source domain dataset, and $\mathcal{S}_t = \{(\bm{x}_j^t, \mathcal{B}_j^t)\}_{j=1}^{N_t}$ denote the limited target domain dataset. 
Here, $\bm{x}_i^s$ and $\bm{x}_j^t$ represent the source and target images, respectively. Nevertheless, the inherent domain shift between synthetic source data ($\mathcal{S}_s$) and real-world target data ($\mathcal{S}_t$), driven by variations in illumination, weather, viewpoints, and resolution, severely limits the performance of the deep learning detector $f_{\bm \Theta}$. In fact, if improperly handled, synthetic data can actively degrade performance~~\autocite{shi2025closer}.  

\medskip
\noindent \textbf{Sim2Real adaptation.}
Consequently, Simulation-to-Real (Sim2Real) adaptation is essential to train a robust detector $f_{\bm \Theta}$ that generalizes to in-the-wild UAV scenarios (e.g., surveillance or crowded scenes).  Existing approaches typically tackle this gap through image- or feature-level adaptation. For instance, Sim2Air~~\autocite{DBLP:journals/ral/BarisicPB22-sim2air} mitigates style and temporal drift via spatiotemporal attention, while Truong et al.~~\autocite{DBLP:journals/ral/TruongCB21-bi-directional} propose a bi-directional framework using dual generative models. For UAV-specific pedestrian detection, Saadiyean et al.~~\autocite{DBLP:conf/icra/SaadiyeanSS24-multi-scale} employ multi-scale adversarial feature alignment. While few works explicitly target UAV human detection, progressive transformation learning (PTL)~~\autocite{shen2023progressive} progressively translates local human regions using a conditional GAN. In contrast to these GAN- or feature-based methods, our framework uniquely leverages diffusion models to execute both global style and local content transformations.

\subsection{Diffusion Models and Image-to-Image Translation}

\noindent \textbf{Basics of diffusion models.}
Diffusion models ~\autocite{zhang2024the,DBLP:conf/nips/HoJA20-ddpm,DBLP:conf/iclr/PodellELBDMPR24,DBLP:conf/cvpr/RombachBLEO22} are powerful generative models that excel in unconditional synthesis, inpainting, and conditional translation ~\autocite{DBLP:conf/nips/HoJA20-ddpm, song2020denoising, yang2024vip, lugmayr2022repaint, DBLP:conf/iclr/MengHSSWZE22-sdedit, chen2024exploring}. Rather than directly estimating the data density $P_0$, they utilize a two-stage stochastic process.  The \textit{forward process} incrementally corrupts data $\bm{x}_0 \sim P_0$ into Gaussian noise, $p(\bm{x}_t \mid \bm{x}_0) = \mathcal{N}\Big(\sqrt{\bar{\alpha}_t}\bm{x}_0,\,(1-\bar{\alpha}_t)\bm{I}\Big)$, which can be formulated continuously as the SDE $d\bm{x} = f(\bm{x},t)\, dt + g(t)\, d\bm{w}$ ~\autocite{DBLP:conf/nips/SongE19}. The corresponding \textit{reverse process} recovers the data via the reverse-time SDE:$$d\bm{x} = \Big[f(\bm{x},t) - g^2(t)\nabla_{\bm{x}} \log p_t(\bm{x})\Big]\, dt + g(t)\, d\bm{w},$$where $dt<0$. To approximate the intractable score function $\nabla_{\bm{x}} \log p_t(\bm{x})$, a neural network $\bm{s}_{\bm \theta}(\bm{x},t)$ is optimized via score matching: $\mathbb{E}_{t, \bm{x}} \Big[ \lambda(t) \big\| \nabla_{\bm{x}} \log p_t(\bm{x}) - \bm{s}_{\bm \theta}(\bm{x}, t) \big\|_2^2 \Big]$. This iterative denoising avoids the training instability and mode collapse of GANs, enabling highly robust image translation. 
BBuilding upon this formulation, Denoising Diffusion Probabilistic Models (DDPM) ~\autocite{DBLP:conf/nips/HoJA20-ddpm} model the reverse diffusion process as a discrete Markov chain that gradually removes noise to recover clean samples.

To improve sampling efficiency and enable controllable editing, Denoising Diffusion Implicit Models (DDIM) ~\autocite{song2020denoising} introduce a deterministic non-Markovian formulation of the reverse process. DDIM preserves the training objective of DDPM while allowing much faster sampling trajectories. Importantly, DDIM admits an approximate \textit{inversion} process that maps a real image into a corresponding latent noise representation along the diffusion trajectory. This property enables image transformation by first encoding an image into the diffusion latent space, perturbing it through controlled DDIM inversion, and then applying guided denoising to generate modified outputs.

In this work, we leverage DDIM inversion enable structured image-to-image transform such as global style transfer (GST) and local refinement (LR) while preserving the semantic structure of the original scene.

\medskip
\noindent \textbf{Image-to-image translation via diffusion models.}  Diffusion models are highly effective for image-to-image (I2I) translation tasks where structural preservation is paramount. For instance, Palette ~\autocite{DBLP:conf/siggraph/SahariaCCLHSF022-palette} tailors conditional diffusion for colorization and inpainting, while ILVR ~\autocite{DBLP:conf/iccv/ChoiKJGY21-ilvr} guides the generative process using low-resolution conditioning to ensure semantic alignment. Similarly, SDEdit ~\autocite{DBLP:conf/iclr/MengHSSWZE22-sdedit} injects spatial conditions like sketches or semantic maps to enable fine-grained, spatially aware synthesis. Finally, techniques like Blended Diffusion ~\autocite{DBLP:conf/cvpr/AvrahamiLF22-blended-diffusion} modulate noise spatially to restrict transformations to specific regions. Collectively, these advancements demonstrate that diffusion models can execute localized, structurally consistent edits—capabilities that are strictly necessary for our localized refinement objectives in Sim2Real transfer.

\subsection{Hierarchical Disentanglement for Domain Adaptation}

Visual domain adaptation traditionally treats the Sim2Real gap as a monolithic transformation, relying on global techniques like adversarial training or feature matching. However, recent research demonstrates that domain shifts occur across multiple semantic levels—such as global style (illumination) and local structure (object geometry)—and that explicitly modeling this hierarchy yields more robust adaptation. For instance, Chen et al.~~\autocite{DBLP:conf/cvpr/Chen0SDG18-domain-adaptation-faster-rcnn} introduced Hierarchical Feature Alignment (HFA) to synchronize both global and class-specific layers, while Hoffman et al.~~\autocite{DBLP:series/acvpr/HoffmanTDS17-simultaneous-deep-transfer} pioneered disentangled alignment by independently addressing inter- and intra-class discrepancies. Building on this, iFAN~~\autocite{DBLP:conf/aaai/ZhuangHHS20-iFAN} effectively integrates image- and instance-level alignments using category-aware modules to resolve multi-level shifts in standard driving datasets.

While these foundational methods successfully explore multi-level feature alignment, they do not utilize generative modeling to resolve domain shifts directly in pixel space. In the following section, we introduce Coarse-to-Fine Hierarchical Alignment (CFHA) framework, which advances this paradigm by employing a dual-diffusion architecture to collaboratively transfer global style and refine local details. This pixel-level hierarchical approach provides the fine-grained, controllable Sim2Real adaptation strictly required for the complex variations inherent to UAV imagery.

\section{Our Method: Coarse-to-Fine Hierarchical Alignment}\label{section_method}

In this work, we introduce Coarse-to-Fine Hierarchical Alignment (CFHA), a novel data augmentation framework designed to transform synthetic data into photorealistic UAV imagery and then utilize these augmented data to train a more powerful detector in specific domain. UAV Sim2Real adaptation presents unique challenges due to high-altitude viewpoints, extremely small object scales, and complex background contexts. We observe that the domain gap between synthetic and real UAV images fundamentally manifests at two distinct levels: (1) \textbf{Global style shifts} (e.g., illumination, color temperature, and rendering artifacts), and (2) \textbf{Local content shifts} (e.g., unnatural human poses and spatial distributions). Attempting to resolve both shifts simultaneously typically yields either over-smoothed global appearances or locally inconsistent artifacts (as shown in \Cref{fig:pipeline}). 

To overcome the challenge, as shown in \Cref{fig:pipeline}, our CFHA explicitly decouples the adaptation process into three hierarchical stages that sequentially aligns global and local shifts followed by local refinement:
\begin{itemize}[leftmargin=*]
\item[1.] \textbf{Stage I: Global Style Transfer (\Cref{subsec:gst})} performs global style alignment in the latent space via diffusion inversion. A structure-aware cross-attention mechanism computes attention weights to extract and fuse aligned style features with content features, yielding a globally styled image that strictly preserves structural fidelity.
\vspace{0.05in}
\item[2.] \textbf{Stage II: Local Refinement (\Cref{subsec:lr})} enhances local fidelity by isolating object-centric patches and generating corresponding text prompts via automated image captioning. Conditioned on these prompts, a fine-tuned diffusion model via LoRA refines the photorealistic quality of each patch.
\vspace{0.05in}
\item[3.] \textbf{Stage III: Hallucination Removal (\Cref{subsec:cr})} eliminates generated artifacts through CLIP-based semantic filtering and mask-guided deletion, guaranteeing clean, data-consistent final outputs.
\end{itemize}
By explicitly disentangling global and local adaptation in the pixel space, our CFHA enables targeted objectives to each stage. This yields translated images that are photorealistic, spatially consistent, and highly optimized for downstream detection tasks.  Ultimately, our final detector $f_{\bm \Theta}$ is trained on the augmented dataset $\tilde{\mathcal{S}}_s \cup \mathcal{S}_t$, where $\tilde{\mathcal{S}}_s$ denotes the transformed synthetic data and $\mathcal{S}_t$ represents the real target data. As detailed in \Cref{sec:exp}, comprehensive empirical studies validate the effectiveness of this hierarchical approach. We utilize the Fréchet Inception Distance (FID)~~\autocite{DBLP:conf/nips/HeuselRUNH17-fid} to quantify the image-level and instance-level quality of our translations and rigorously evaluate the resulting detection performance on standard UAV benchmarks. 

Crucially, our ablation studies in \Cref{sec:exp} reveal a strong synergy across all three stages; deploying any single stage or partial combination yields only marginal improvements, underscoring the necessity of our unified framework. Next, we introduce each stage in detail.

\begin{figure*}[!t]
    \centering
    \includegraphics[width=\linewidth]{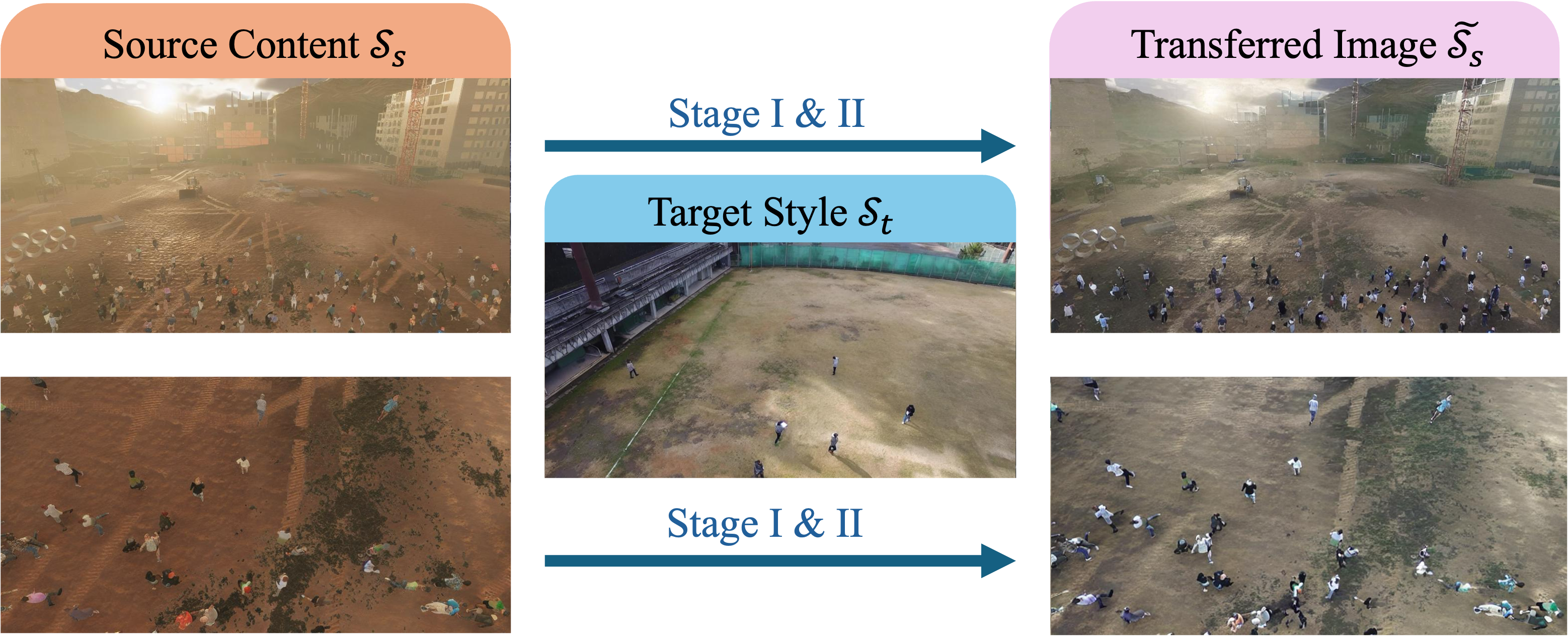} \\
    \caption{Visualization of transformation on Okutama-Action with only Stage I and Stage II combined.}
    \vspace{-0.2in}
    \label{fig:three-steps_styletransfer}
\end{figure*}

\subsection{Stage I: Global Style Transfer (GST)}
\label{subsec:gst}

Motivated by the insight that domain adaptation is inherently hierarchical, the first stage of our CFHA executes a training-free global style transfer via latent diffusion models. This stage explicitly aligns macroscopic appearance characteristics—such as color palette, illumination, and atmospheric haze—between a synthetic content image $\bm{x}_i^s$ and a small set of real reference images, all while strictly preserving spatial labels.  Inspired by recent advancements~~\autocite{DBLP:journals/corr/abs-2505-16360-cactif}, we achieve this alignment by adopting a streamlined \emph{image-wise} Adaptive Instance Normalization (AdaIN) mechanism directly within the latent space.

Specifically,  let $(\mathcal E,\mathcal D)$ be the VAE encoder/decoder of the Stable‐Diffusion
backbone. Encoding a content image $\bm{x}_i^s$ and a style example $\bm{x}_j^t$ yields latents
$\bm{z}_{i,0}^s=\mathcal E(\bm{x}_i^s)$ and $\bm{z}_{j,0}^t=\mathcal E(\bm{x}_j^t)$.  
After DDIM inversion to timestep $T$, we obtain noisy latents
$\bm{z}_{i,T}^s, \bm{z}_{j,T}^t \in\mathbb{R}^{C'\times H'\times W'}$.
For each reverse step $t=T,\dots,1$ we perform

\begin{equation}
\bm{z}_{i,t-1} \;=\;
\mathrm{UNet}_{\bm \phi}
\Bigl(\,
\underbrace{\text{AdaIN}(\hat{\bm{z}}^s_{i,t}, \bm{z}^t_{i,t})}_{\text{global statistics}},
\,t\Bigr),
\label{eq:gst_step}
\end{equation}
where we have
\begin{align}
\bm Q &= \bm W_q^s \bm{z}^s_{i,t}, \quad \bm K = \bm W_k^t \bm{z}^t_{j,t},\quad \bm V = \bm W_v^t \bm{z}^t_{j,t}, \\
\bm A &= \text{softmax} \left( \frac{\bm Q \cdot \bm K^\top}{\sqrt{d}} \right), \hat{\bm{z}}^s_{i,t} = \bm A \cdot \bm V \label{eq:cross_attn}, \\
\text{AdaIN}&(\hat{\bm{z}}^s_{i,t}, \bm{z}^t_{i,t}) 
= \sigma(\bm{z}^t_{i,t}) \cdot \frac{\hat{\bm{z}}^s_{i,t} - \mu(\hat{\bm{z}}^s_{i,t})}{\sigma(\hat{\bm{z}}^s_{i,t})} + \mu(\bm{z}^t_{i,t}). \label{eq:adaIN}
\end{align}
where $\mu(\cdot)$ and $\sigma(\cdot)$ denote the channel-wise mean and standard deviation computed over spatial dimensions.
As such, our GST aligns macroscopic appearance statistics—such as color temperature, illumination, and atmospheric haze—while strictly preserving the spatial structure of the synthetic source image. After $T$ reverse steps, the decoder produces $\tilde{\bm{x}}=\mathcal D(\bm{z}_0)$, a globally photorealistic image that closely matches the target distribution. Crucially, the underlying object geometry remains unchanged, ensuring the absolute validity of all original bounding-box annotations.  These globally styled images subsequently serve as the input for the Local Refinement stage in \Cref{subsec:lr}, which systematically enhances local instance-level quality and restores fine-grained semantic details.

\subsection{Stage II: Local Refinement (LR)}
\label{subsec:lr}
The second stage targets the \emph{local shifts} left unresolved by \text{Global Style Transfer}: GST often distorts small-scale human instances,
which occupy only a few pixels in UAV imagery. Blurred object contours, over-smoothed textures, and human
artifacts (see \Cref{fig:pipeline}) are easily observed. We achieve this by employing a \textbf{super-resolution} diffusion model that up-scales GST outputs by a factor~$Scale{=}2$ to add fine-grained details to local instance level objects.

Specifically, inspired by OSEDiff~~\autocite{DBLP:conf/nips/WuS0Z24-OSEDiff}, we treat the GST output $\tilde{\bm{x}} \in \mathbb{R}^{H\times W\times3}$ as a \emph{latent starting point} rather than starting from a pure random noise. To adapt the pre-trained Stable Diffusion backbone in Stage I for the super-resolution task, we apply few-shot LoRA fine-tuning to the encoder-decoder pair $(\mathcal{E}, \mathcal{D})$, yielding the refined modules $\mathcal{E}'$ and $\mathcal{D}'$. We apply an identical LoRA fine-tuning strategy to the latent diffusion network $\bm{\epsilon}_{\bm{\theta}}$.  During inference, given an upsampling factor of $Scale=2$, we first generate a bicubic upsampled image $\bm{x}_{\uparrow}$. This is encoded into the latent space as $\bm{z}_L = \mathcal{E}'(\bm{x}_{\uparrow})$, which serves as the initialization for a single reverse diffusion step at timestep $T$: 
\begin{gather}
\bm{z}_H = F_{\bm \theta}(\bm{z}_L;\,\bm{c}_y)
    = \frac{\bm{z}_L - \beta_T\, \bm {\epsilon}_{\bm \theta}(\bm{z}_L;T,\bm{c}_y)}{\alpha_T}
    \label{eq:lr_onestep_a} \\
\hat{\bm{x}} = \mathcal D'(\bm{z}_H),
    \label{eq:lr_onestep_b}
\end{gather}
    where $(\alpha_T,\beta_T)$ are DDPM coefficients, $\bm{c}_y$ is a tag-style prompt extracted from $\tilde{\bm{x}}$ via DAPE~~\autocite{wu2024seesr,DBLP:conf/nips/WuS0Z24-OSEDiff}, and $\bm{z}_L$ is only modified within the predefined mask corresponding to each instance. Additional training details are provided in \Cref{sec:stage2training}.

Our training minimizes a weighted data term and a latent variational score distillation (VSD) loss that aligns the predicted noise $\bm{\epsilon}_{\bm{\theta}}$ with that of the frozen teacher $\bm{\epsilon}_{\bm{\theta}'}$. As such, our one-step mechanism in \eqref{eq:lr_onestep_a} restores fine textures while avoiding the high cost of multi-step sampling. We visualize examples of style transformation in \Cref{fig:pipeline} and \Cref{fig:three-steps_styletransfer}.

\subsection{Stage III: Hallucination Removal (HR)}
\label{subsec:cr}
\vspace{4pt}

\begin{figure*}[t]
    \centering
    \includegraphics[width=\linewidth]{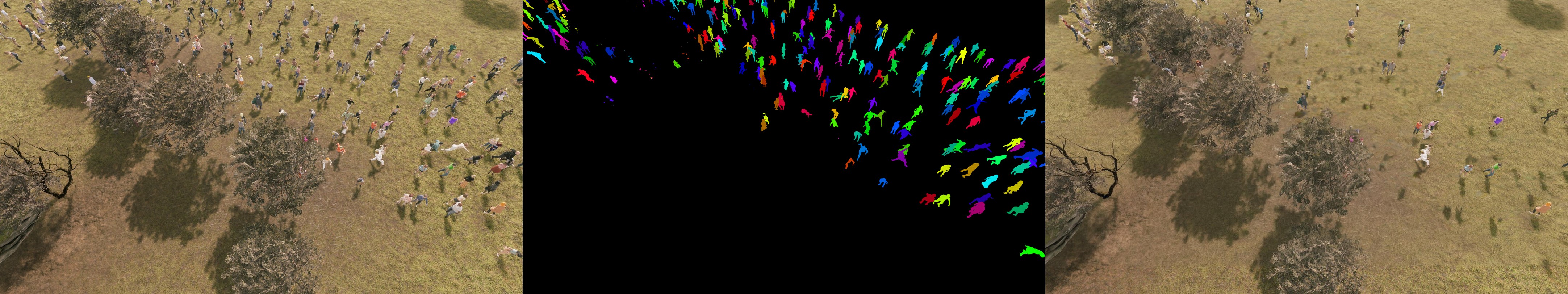} \\
    \includegraphics[width=\linewidth]{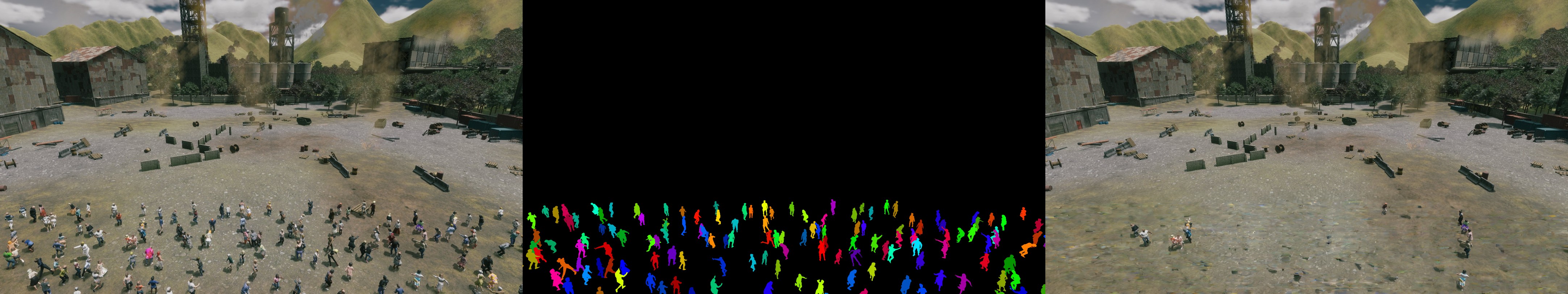} 
    \caption{\textbf{Visualization of Hallucination Removal (HR) on Okutama.} \underline{Left Column:} Original synthetic data, \underline{Middle Column:} mask annotations of the images, \underline{Right Column:} Transferred data after HR.}
    \label{fig:three-steps_removal}
\end{figure*}

Even after Local Refinement, the generative pipeline may still produce hallucinated or structurally unrealistic human instances. As shown in \Cref{fig:three-steps_removal}, to actively suppress these artifacts, we introduce a \textbf{CLIP-guided similarity filter} designed to retain only those instances whose visual distributions closely align with real-world data.  We leverage a pre-trained CLIP model~~\autocite{DBLP:conf/icml/RadfordKHRGASAM21} for its robust, web-scale generalization. Crucially, its joint image-text embedding space allows us to use precise textual prompts as semantic anchors, ensuring the filtered visual concepts remain strictly faithful to the target domain.

To circumvent the computational bottleneck of a nearest-neighbor search against a massive memory bank of real human embeddings $\{\bm{u}_m\}$, we distill their distribution into a single, prototypical reference vector $\bm{t}_*$. This vector is optimized to maximize similarity with the mean embedding of the real examples, $\bar{\bm{u}} = \text{normalize}(\frac{1}{M}\sum_m \bm{u}_m)$, while remaining anchored to a text embedding, $\bm{t}_{\text{anchor}}$, generated from a domain-specific prompt (e.g., ``a photo of a person taken from a drone'').  Acting as a semantic regularizer, this text anchor grounds the visual prototype. Formally, we seek a vector $\bm{v}$ on the unit hypersphere that maximizes the objective:$$\mathcal{L}(\bm{v}) = \bm{v}^\top \bar{\bm{u}} + \lambda ( \bm{v}^\top \bm{t}_{\text{anchor}} ) \label{eq:loss_func}$$where $\lambda$ controls the regularization strength of the text anchor. Rather than relying on iterative gradient descent, this objective admits an efficient closed-form solution. The optimal reference vector $\bm{t}_*$ is obtained by simply normalizing the weighted sum:$$\bm{t}_* = \frac{\bar{\bm{u}} + \lambda \, \bm{t}_{\text{anchor}}}{\|\bar{\bm{u}} + \lambda \, \bm{t}_{\text{anchor}}\|_2} \label{eq:closed_form}$$Utilizing this learned reference, we project each synthetic person crop $\bm{c}_k$ into the embedding space via $\bm{v}_k = \mathrm{CLIP}(\bm{c}_k)$, defining its similarity score $s_k$ as the cosine similarity to $\bm{t}_*$. These raw scores are subsequently mapped into a sampling distribution via a softmax function:$$s_k = \frac{\bm{v}_k^{\!\top}\bm{t}_*} {\|\bm{v}_k\|_2\|\bm{t}_*\|_2}, \qquad p_k = \frac{\exp(\tau s_k)}{\sum_j \exp(\tau s_j)}, \label{eq:clip_score}$$where the temperature hyperparameter $\tau$ modulates the sharpness of the distribution.  During mini-batch construction, we \emph{stochastically retain} each crop $k$ with probability $p_k$. All unselected instances are systematically masked from the images and excluded from the annotations, yielding a rigorously cleaned training set. We visualize examples of this Hallucination Removal (HR) process in \Cref{fig:three-steps_removal}.

\section{Experiments}\label{sec:exp}

In this section, we present comprehensive experiments demonstrating the efficacy of our proposed framework in mitigating both style and content domain gaps across multiple datasets. We begin by highlighting its performance in detection accuracy in a cross-domain setting.

\subsection{Experimental Setup}

\noindent \textbf{Evaluation datasets.} Our adaptation is performed on the synthetic UAV datasets and is expected to improve the performance of the detector on real UAV datasets. Since the annotated target images are very sparse in the real application, the training dataset contains only $20$ real images from the target domain and $500$ synthetic images from the source domain.

We use the \textbf{SynPlay} dataset ~\autocite{yim2024synplay} as the synthetic dataset. \textbf{SynPlay} is a synthetic human dataset featuring over 73k images and 6.5M human instances, designed to capture diverse human appearances with realistic motions and multiple camera viewpoints. It enables improved human detection and segmentation, particularly for data-scarce and cross-domain learning scenarios.

For the real datasets, we adopt four benchmarks, Okutama-Action ~\autocite{DBLP:conf/cvpr/BarekatainMSMNM17}, Semantic drone ~\autocite{Fraundorfer2019DroneDataset-semantic-drone}, VisDrone2018 ~\autocite{DBLP:journals/corr/abs-1804-07437} and Manipal-UAV ~\autocite{akshatha2023manipal}. 
\begin{itemize}[leftmargin=*]
    \item[1.] \textbf{Okutama-Action} ~\autocite{DBLP:conf/cvpr/BarekatainMSMNM17} is a video dataset for aerial-view concurrent human action detection, containing 43 fully-annotated minute-long sequences with 12 action classes. It introduces challenges such as dynamic action transitions, large-scale and aspect ratio changes, abrupt camera movements, and multi-labeled actors, making it more realistic and difficult than existing datasets.
    \vspace{0.03in}
    \item[2.] \textbf{Semantic Drone} ~\autocite{Fraundorfer2019DroneDataset-semantic-drone} provides high-resolution urban aerial imagery for advancing autonomous drone perception. It includes 600 images with pixel-accurate annotations for 20 semantic classes and bounding boxes for person detection, supporting both object detection and semantic segmentation tasks. 
        \vspace{0.03in}
    \item[3.] \textbf{VisDrone2018} ~\autocite{DBLP:journals/corr/abs-1804-07437} offers 263 videos and 10,209 images with over 2.5 million annotated instances for object detection and tracking in diverse drone scenes.
        \vspace{0.03in}
    \item[4.] \textbf{Manipal-UAV-Person}~~\autocite{akshatha2023manipal} is a dataset that comprises 33 UAV videos and 13,462 sampled images with 153,112 annotated person instances. The data was captured using multiple drone platforms under diverse altitudes, lighting, and weather conditions, and is specifically designed to benchmark small-scale and densely distributed human detection in challenging UAV scenarios.
\end{itemize}

 
\noindent \textbf{Evaluation metrics.} In our evaluation for the detector, we report the mAP@50 and mAP@50-95: (i) \textbf{mAP@50} (mean Average Precision at an IoU threshold of 0.5) evaluates the average detection performance under a relatively loose matching criterion. (ii) \textbf{mAP@50-95}, following the COCO evaluation protocol, averages mAP across IoU thresholds from 0.5 to 0.95 with a step size of 0.05, providing a more comprehensive and stringent assessment of detection accuracy under varying overlap conditions.

\subsection{Implementation Details of CFHA}
Our framework is implemented in PyTorch and built upon the Stable Diffusion 2.1 backbone~~\autocite{rombach2022high}. 
Stage I performs global style transfer via attention-modulated latent transformations to align synthetic and real UAV imagery. 
Stage II conducts local refinement by adapting the diffusion backbone with a lightweight LoRA module~~\autocite{hu2022lora}, trained on degraded real instances and patches following the OSEDiff objective~~\autocite{DBLP:conf/nips/WuS0Z24-OSEDiff}. 
Stage III removes hallucinated artifacts through an attentive diffusion-based erasing mechanism~~\autocite{cui2025attentiveeraserunleashingdiffusion}. 
To further enforce semantic consistency, we employ a CLIP-guided similarity filter using the pre-trained ViT-L/14 model from OpenCLIP~~\autocite{fang2023datafilteringnetworks,ilharco_gabriel_2021_5143773}.

During our experiments, the training set comprises 500 synthetic content images from SynPlay~~\autocite{yim2024synplay} and a few-shot set of 20 real reference images sampled from one of the four target UAV datasets. Finally, for the CLIP-guided filtering, we set the text-anchor regularization weight to $\lambda=0.2$, strictly grounding the generated visual prototype within the desired semantic context.

\medskip 
\noindent \textbf{Training object detectors.}
Once the CFHA framework generates the adapted synthetic dataset $\tilde{\mathcal{S}}_s=\{(\tilde{\bm{x}}_i^s,\mathcal{B}_i^s)\}$, we jointly train the object detector $f_{\bm \Theta}$ using both the real target images $\mathcal{S}_t$ and the translated images $\tilde{\mathcal{S}}_s$.  This dual training paradigm fundamentally stabilizes the learning process: the original target data guarantees exact geometric fidelity, while the translated synthetic data injects scalable appearance realism. The total detection loss is formulated as:
\[\begin{aligned} \mathcal{L}_{\text{det-total}}(\bm \Theta) = &\; \lambda_{\text{orig}} \, \mathbb{E}_{(\bm{x},\mathcal{B})\in\mathcal{S}_t} \bigl[\mathcal{L}_{\text{det}}(f_{\bm \Theta}(\bm{x}),\mathcal{B})\bigr] + \lambda_{\text{tran}} \, \mathbb{E}_{(\tilde{\bm{x}},\mathcal{B})\in\tilde{\mathcal{S}}_s} \bigl[\mathcal{L}_{\text{det}}(f_{\bm \Theta}(\tilde{\bm{x}}),\mathcal{B})\bigr], \end{aligned}\]
where $\mathcal{L}_{\text{det}}$ represents the standard detection objective (comprising classification and bounding-box regression), and $\lambda_{\text{orig}}, \lambda_{\text{tran}} > 0$ are scalar weighting factors. For our experiments, we deploy YOLOv11x as the core detection architecture. While $\lambda_{\text{orig}}$ and $\lambda_{\text{tran}}$ can be dynamically tuned to mitigate imbalances between the synthetic and real data distributions, we empirically set both to $1$ to equally weight both domains during optimization.

\subsection{Experimental Results}

To validate the effectiveness of our Sim2Real adaptation framework, we train human detectors using either the original synthetic images or the CFHA-transformed images and compare their performance on real-world UAV benchmarks. As shown in \Cref{tab:augmentation_results}, CFHA consistently improves detection performance across all four evaluated datasets. 

In particular, on the Semantic-Drone dataset, CFHA achieves a substantial improvement over the synthetic baseline, boosting mAP@50 from 18.6 to 32.7 (\textbf{+14.1}) and mAP@50–95 from 8.3 to 15.0 (\textbf{+6.7}). Similar gains are observed on the Okutama dataset, where CFHA improves mAP@50 from 64.6 to 68.2 and mAP@50–95 from 22.4 to 25.3. CFHA also achieves the best performance on the Manipal-UAV and VisDrone benchmark. 

\begin{figure}[t!]
  \centering
  \begin{subfigure}[t]{0.48\columnwidth}
    \centering
    \includegraphics[width=\linewidth]{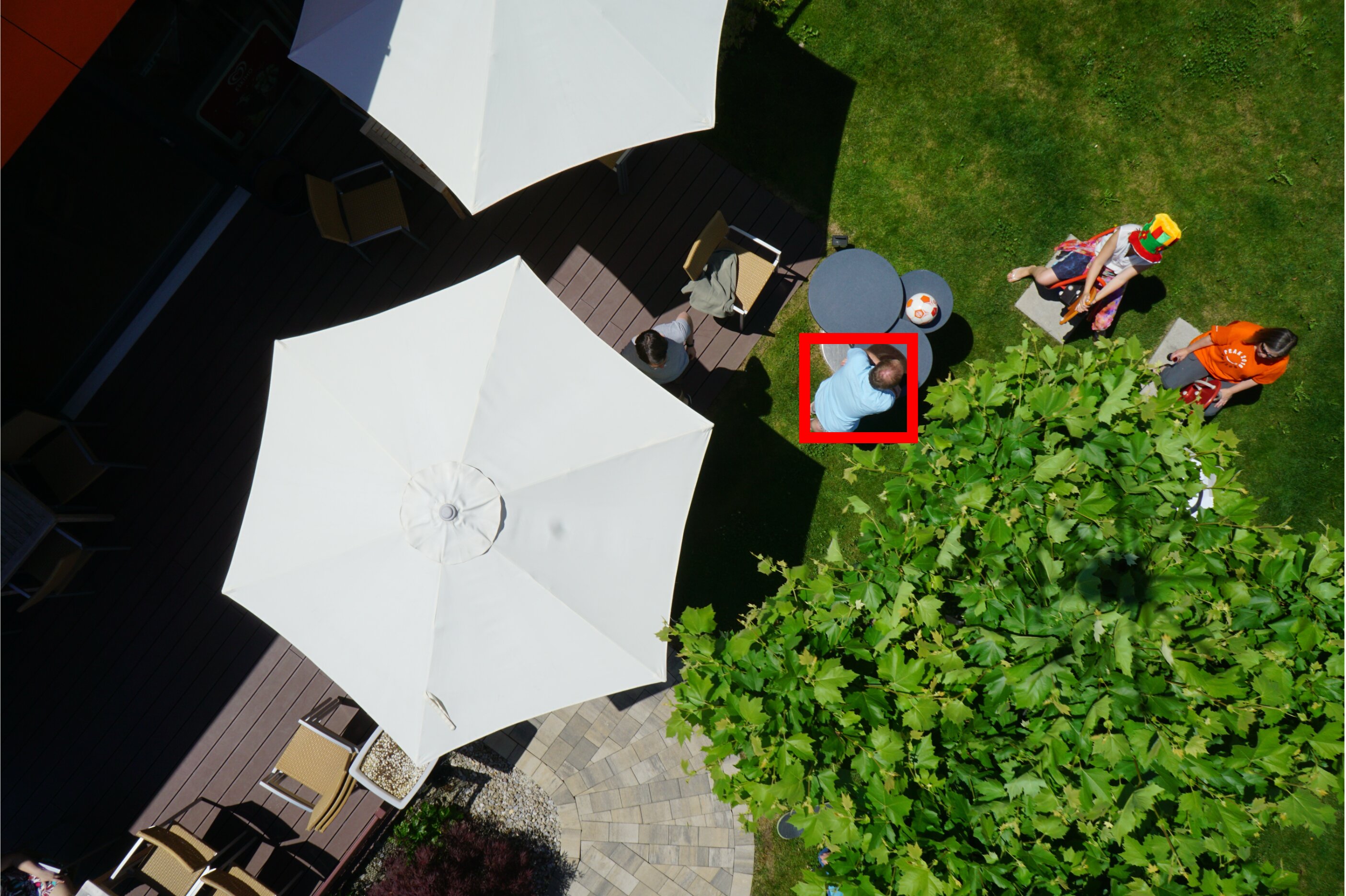}
    \caption{}
    \label{fig:det-baselinede}
  \end{subfigure}\hfill
  \begin{subfigure}[t]{0.48\columnwidth}
    \centering
    \includegraphics[width=\linewidth]{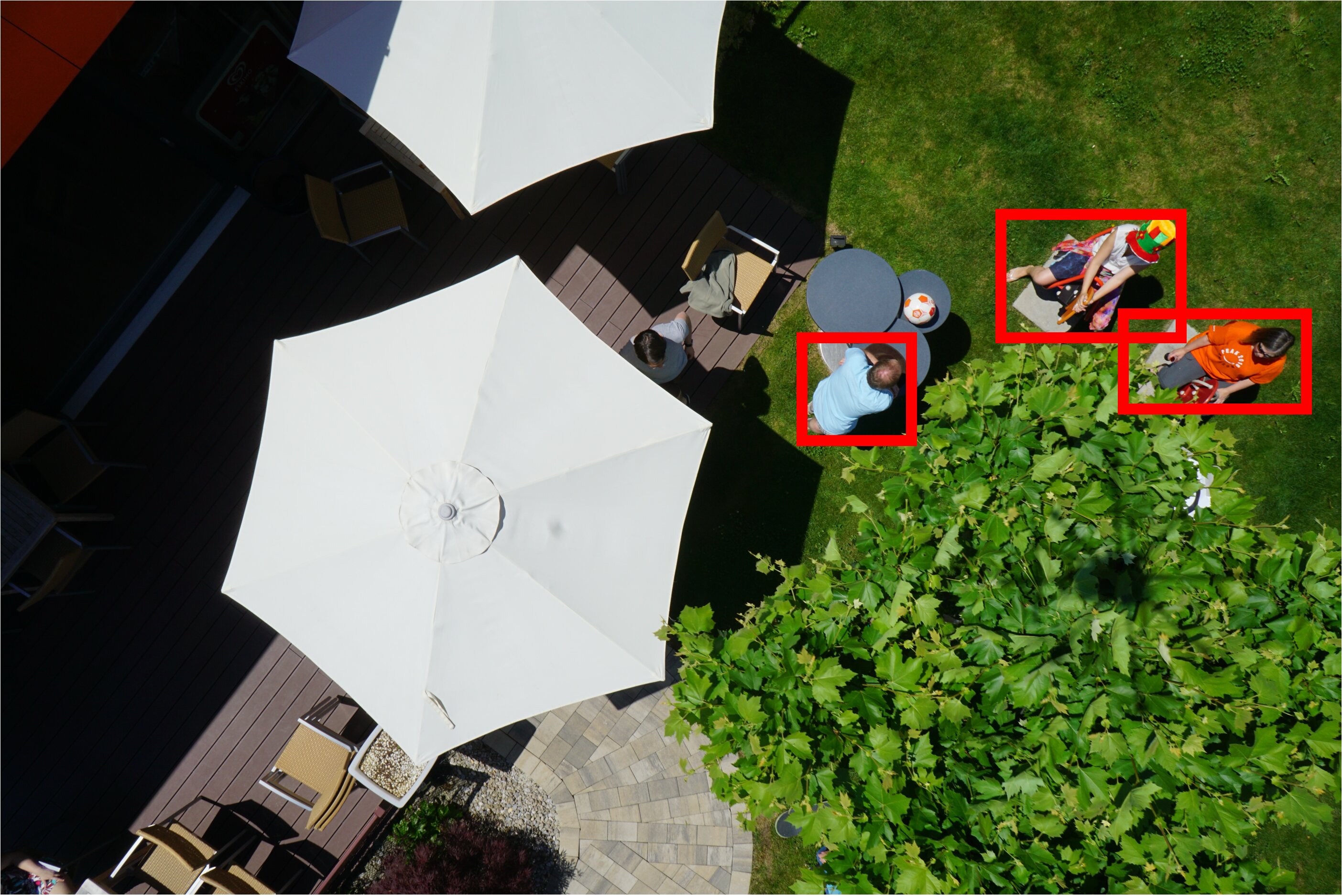}
    \caption{}
    \label{fig:det-ours}
  \end{subfigure}
  \caption{\textbf{Human detection results comparison of Semantic Drone dataset.} 
  (a) Baseline: Detection results trained on \textbf{original Synplay} data.
  (b) Ours: Detection results trained on \textbf{CHFA}-transformed Synplay data.}
  \label{fig:det-comparison}
\end{figure}

On the VisDrone dataset, the improvements are comparatively modest. This can be attributed to the sparse ground-truth annotations in the VisDrone test set, where a large proportion of images contain no human instances, making human-centric evaluation metrics inherently unstable. Qualitative comparisons of the resulting detections are provided in \Cref{fig:det-comparison}.

\begin{table}[t]
\centering
\resizebox{\textwidth}{!}{%
\begin{tabular}{l|cc|cc|cc|cc}
\toprule
\multirow{2}{*}{\textbf{Method}} & \multicolumn{2}{c|}{\textbf{Okutama}} & \multicolumn{2}{c|}{\textbf{Semantic-Drone}} & \multicolumn{2}{c|}{\textbf{VisDrone}} & \multicolumn{2}{c}{\textbf{Manipal-UAV}} \\
\cmidrule(lr){2-3} \cmidrule(lr){4-5} \cmidrule(lr){6-7} \cmidrule(lr){8-9} & \textbf{mAP50} & \textbf{mAP95} & \textbf{mAP50} & \textbf{mAP95} & \textbf{mAP50} & \textbf{mAP95} & \textbf{mAP50} & \textbf{mAP95} \\
\midrule

Synthetic (baseline) & 64.60 & 22.40 & 18.60 & 8.30 & 25.90 & 11.10 & 47.80 & 18.30 \\

\midrule
CUT ~\autocite{zhang2019multimodal}                  & 41.21 & 16.22 & 14.26 & 5.83 & 19.93 & 14.68 & 50.23 & 19.58 \\
IP-Adapter ~\autocite{ye2023ip}           & 56.32 & 21.85 & 17.76 & 9.45 & 24.83 & 13.11 & 51.35 & 19.24 \\
PTL ~\autocite{shen2023progressive}                  & 65.74 & 24.43 & 27.38 & 13.79 & 28.58 & 12.97 & 50.26 & 19.73 \\
\textbf{Ours (CFHA)}  & \textbf{68.20} & \textbf{25.30} & \textbf{32.70} & \textbf{15.00} & \textbf{31.50} & \textbf{15.10} & \textbf{53.80} & \textbf{20.50} \\

\bottomrule
\end{tabular}}
\vspace{0.05in}
\caption{\textbf{Performance comparison of synthetic data augmentation methods on four aerial-view benchmarks.}}
\label{tab:augmentation_results}
\end{table}

\subsection{Ablation Study}

In this subsection, we conduct comprehensive ablation studies to isolate and evaluate the individual contributions of each stage within the CFHA framework. As detailed in \Cref{section_method}, our pipeline comprises Global Style Transfer (GST), Local Refinement (LR), and Hallucination Removal (HR), which collaborate to align both style and content.  As shown in \Cref{tab:ablation1}, our step-wise analysis reveals the following:
\begin{itemize}[leftmargin=*]
\item \textbf{GST alone} actively degrades detection performance, as the global translation tends to introduce destructive artifacts onto small human instances.
\item \textbf{LR} successfully recovers and enhances detection performance by aggressively eliminating these localized artifacts and injecting structural realism.
\item \textbf{HR} further boosts detector accuracy by actively filtering out structurally anomalous instances, perfectly aligning the content distributions.
\end{itemize}

Furthermore, we investigate the FID metrics (patch-wise vs. image-wise) shown in \cref{sec:FID}.  As shown in \Cref{tab:fid_ablation}, merely reducing the global background discrepancy (i.e., lowering image-wise FID from 320 to 235 via GST) paradoxically degrades mAP@50. This is because GST slightly worsens the patch-wise FID (107 vs. 104), confirming that global alignment corrupts local human regions.

\begin{table}[t]
  \centering
  \resizebox{\textwidth}{!}{%
    \begin{tabular}{l|cc|cc|cc|cc}
      \toprule
      \multirow{2}{*}{\textbf{Variants}} & \multicolumn{2}{c|}{\textbf{Okutama}} & \multicolumn{2}{c|}{\textbf{Semantic-Drone}} & \multicolumn{2}{c|}{\textbf{VisDrone}} & \multicolumn{2}{c}{\textbf{Manipal-UAV}} \\
      \cmidrule(lr){2-3} \cmidrule(lr){4-5} \cmidrule(lr){6-7} \cmidrule(lr){8-9}
       & \textbf{mAP50} & \textbf{mAP50-95} & \textbf{mAP50} & \textbf{mAP50-95} & \textbf{mAP50} & \textbf{mAP50-95} & \textbf{mAP50} & \textbf{mAP50-95} \\
      \midrule
      Synthetic (Baseline) & 64.6 & 22.4 & 18.6 & 8.3  & 25.2 & 11.1 & 47.8 & 18.3 \\
      \midrule
      GST only             & 30.3 & 8.8  & 13.9 & 5.6  & 23.5  & 10.6  & 51.36   &  20.8  \\
      LR only             & 38.6 & 10.3  & 19.2 & 8.4  & \textbf{27.3}  & \textbf{11.7}  &  \textbf{57.8}  &  \textbf{23.4}  \\
      HR only             & 35.4 & 9.4  & 20.7 & 9.4  & 21.3  & 8.8  & 52.8   & 19.2   \\
      GST + LR (w/o HR)    & 67.8 & 25.1 & 21.2 & 10.2  & 23.3 & 10.41 & 50.8   & 20.2   \\
      \textbf{GST + LR + HR (Ours)} & \textbf{68.2} & \textbf{25.3} & \textbf{32.7} & \textbf{15.0} & \textbf{31.5} & \textbf{15.1} & \textbf{53.8} & \textbf{20.5} \\
      \bottomrule
    \end{tabular}%
  }
  \vspace{0.05in}
\caption{\textbf{Ablation study.} Detection performance of methods incorporating different components of the proposed CFHA framework, evaluated on four aerial-view datasets.}
\label{tab:ablation1}
\end{table}

However, integrating Local Refinement (``GST + LR") triggers a massive drop in patch-wise FID (from 107 to 69), despite causing the image-wise FID to increase (from 235 to 323). Because LR specifically utilizes caption-guided diffusion to enhance local instance patches while freezing the background, it creates a slight stylistic inconsistency between the hyper-realistic foreground and the globally styled background—driving up the overall image FID. Yet, this localized realism yields substantial detection gains under the Okutama datasets, surging from 64.6 to 68.2 in mAP@50 and 22.4 to 25.3 in mAP@50--95. Ultimately, these results prove that \emph{patch-level realism} around object instances is a vastly superior predictor of downstream detection performance than global visual quality, empirically validating our motivation for an object-centric, multi-stage refinement pipeline.

\section{Conclusion}
We introduced \textbf{CFHA}, a hierarchical diffusion-based framework that bridges both the global style gap and the local content gap for UAV human detection. By combining global style transfer with super-resolution refinement and local hallucination removal, CFHA effectively narrows the Sim2Real domain gap while preserving annotation fidelity. Experiments on public benchmarks show that CFHA significantly improves detection performance, achieving up to \textbf{+14.1} mAP@{50} over baselines. Our results highlight the importance of jointly aligning style and content for UAV human detection and provide an effective method to improve performance in human detection.

\section*{Acknowledgement}
This research was supported by the Army Research Office (ARO) under Grant No. ARO W911NF-25-1-0047. We also acknowledge funding support from ONR N000142512339, DARPA HR0011-25-2-0042, and Google Research Scholar Award.

\printbibliography

@String(AAAI  = {AAAI})

@inproceedings{wang2024convolution,
  title={Convolution meets transformer: Efficient hybrid transformer for semantic segmentation with very high resolution imagery},
  author={Wang, Yuji and Zhao, Ruojun and Wei, Shicai and Ni, Jingchen and Wu, Meng and Luo, Yang and Luo, Chunbo},
  booktitle={IEEE International Geoscience and Remote Sensing Symposium},
  year={2024},
}

@article{shi2025closer,
  title={A Closer Look at Model Collapse: From a Generalization-to-Memorization Perspective},
  author={Shi, Lianghe and Wu, Meng and Zhang, Huijie and Zhang, Zekai and Tao, Molei and Qu, Qing},
  journal={Advances in Neural Information Processing Systems},
  year={2025}
}

@inproceedings{DBLP:conf/cvpr/BarekatainMSMNM17,
  author       = {Mohammadamin Barekatain and
                  Miquel Mart{\'{\i}} and
                  Hsueh{-}Fu Shih and
                  Samuel Murray and
                  Kotaro Nakayama and
                  Yutaka Matsuo and
                  Helmut Prendinger},
  title        = {Okutama-Action: An Aerial View Video Dataset for Concurrent Human Action Detection},
  booktitle    = {IEEE/CVF Conference on Computer Vision and Pattern Recognition Workshops},
    year       = {2017},
}

@inproceedings{DBLP:conf/nips/SongE19,
  author       = {Yang Song and
                  Stefano Ermon},
  title        = {Generative Modeling by Estimating Gradients of the Data Distribution},
  booktitle    = {Advances in Neural Information Processing Systems},
  year         = {2019},
}

@article{DBLP:journals/corr/abs-1804-07437,
  author       = {Pengfei Zhu and
                  Longyin Wen and
                  Xiao Bian and
                  Haibin Ling and
                  Qinghua Hu},
  title        = {Vision Meets Drones: A Challenge},
  journal      = {arXiv preprint arXiv:1804.07437},
  year         = {2018},
}

@software{ilharco_gabriel_2021_5143773,
  author       = {Ilharco, Gabriel and
                  Wortsman, Mitchell and
                  Wightman, Ross and
                  Gordon, Cade and
                  Carlini, Nicholas and
                  Taori, Rohan and
                  Dave, Achal and
                  Shankar, Vaishaal and
                  Namkoong, Hongseok and
                  Miller, John and
                  Hajishirzi, Hannaneh and
                  Farhadi, Ali and
                  Schmidt, Ludwig},
  title        = {OpenCLIP},
  year         = 2021,
}

@misc{fang2023datafilteringnetworks,
      title={Data Filtering Networks}, 
      author={Alex Fang and Albin Madappally Jose and Amit Jain and Ludwig Schmidt and Alexander Toshev and Vaishaal Shankar},
      year={2023},
      eprint={2309.17425},
      archivePrefix={arXiv},
}

@misc{cui2025attentiveeraserunleashingdiffusion,
      title={Attentive Eraser: Unleashing Diffusion Model's Object Removal Potential via Self-Attention Redirection Guidance}, 
      author={Benlei Cui and Wenhao Sun and Xue-Mei Dong and Jingqun Tang and Yi Liu},
      year={2025},
      eprint={2412.12974},
      archivePrefix={arXiv},
}

@inproceedings{DBLP:conf/eccv/LinMBHPRDZ14,
  author       = {Tsung{-}Yi Lin and
                  Michael Maire and
                  Serge J. Belongie and
                  James Hays and
                  Pietro Perona and
                  Deva Ramanan and
                  Piotr Doll{\'{a}}r and
                  C. Lawrence Zitnick},
  title        = {Microsoft {COCO:} Common Objects in Context},
  booktitle    = {European Conference on Computer Vision},
  year         = {2014},

}

@article{yim2024synplay,
  title={SynPlay: Importing Real-world Diversity for a Synthetic Human Dataset},
  author={Yim, Jinsub and Lee, Hyungtae and Eum, Sungmin and Shen, Yi-Ting and Zhang, Yan and Kwon, Heesung and Bhattacharyya, Shuvra S},
  journal={arXiv preprint arXiv:2408.11814},
  year={2024}
}

@inproceedings{DBLP:conf/cvpr/TremblayPABJATC18,
  author       = {Jonathan Tremblay and
                  Aayush Prakash and
                  David Acuna and
                  Mark Brophy and
                  Varun Jampani and
                  Cem Anil and
                  Thang To and
                  Eric Cameracci and
                  Shaad Boochoon and
                  Stan Birchfield},
  title        = {Training Deep Networks With Synthetic Data: Bridging the Reality Gap by Domain Randomization},
  booktitle    = {IEEE/CVF Conference on Computer Vision and Pattern Recognition Workshops},
  year         = {2018},
}

@inproceedings{DBLP:conf/nips/FuTSC0DI23,
  author       = {Stephanie Fu and
                  Netanel Tamir and
                  Shobhita Sundaram and
                  Lucy Chai and
                  Richard Zhang and
                  Tali Dekel and
                  Phillip Isola},
  title        = {DreamSim: Learning New Dimensions of Human Visual Similarity using Synthetic Data},
  booktitle    = {Advances in Neural Information Processing Systems},
  year         = {2023},
}

@article{DBLP:journals/corr/abs-1806-09755,
  author       = {Xingchao Peng and
                  Ben Usman and
                  Kuniaki Saito and
                  Neela Kaushik and
                  Judy Hoffman and
                  Kate Saenko},
  title        = {Syn2Real: {A} New Benchmark forSynthetic-to-Real Visual Domain Adaptation},
  journal      = {CoRR},
  volume       = {abs/1806.09755},
  year         = {2018},
}

@inproceedings{DBLP:conf/cvpr/RombachBLEO22,
  author       = {Robin Rombach and
                  Andreas Blattmann and
                  Dominik Lorenz and
                  Patrick Esser and
                  Bj{\"{o}}rn Ommer},
  title        = {High-Resolution Image Synthesis with Latent Diffusion Models},
  booktitle    = {IEEE/CVF Conference on Computer Vision and Pattern Recognition},
  year         = {2022},
}

@inproceedings{DBLP:conf/iclr/PodellELBDMPR24,
  author       = {Dustin Podell and
                  Zion English and
                  Kyle Lacey and
                  Andreas Blattmann and
                  Tim Dockhorn and
                  Jonas M{\"{u}}ller and
                  Joe Penna and
                  Robin Rombach},
  title        = {{SDXL:} Improving Latent Diffusion Models for High-Resolution Image
                  Synthesis},
  booktitle    = {International Conference on Learning Representations},
  year         = {2024},
}

@article{DBLP:journals/ral/BarisicPB22-sim2air,
  author       = {Antonella Barisic and
                  Frano Petric and
                  Stjepan Bogdan},
  title        = {Sim2Air - Synthetic Aerial Dataset for {UAV} Monitoring},
  journal      = {{IEEE} Robotics and Automation Letter},
  year         = {2022},
}

@article{DBLP:journals/ral/TruongCB21-bi-directional,
  author       = {Joanne Truong and
                  Sonia Chernova and
                  Dhruv Batra},
  title        = {Bi-Directional Domain Adaptation for Sim2Real Transfer of Embodied
                  Navigation Agents},
  journal      = {{IEEE} Robotics Automation Letter},
  year         = {2021},
}

@inproceedings{DBLP:conf/icra/SaadiyeanSS24-multi-scale,
  author       = {Qiranul Saadiyean and
                  S. P. Samprithi and
                  Suresh Sundaram},
  title        = {Learning Multi-Scale Context Mask-RCNN Network for Slant Angled Aerial
                  Imagery in Instance Segmentation in a Sim2Real setup},
  booktitle    = {{IEEE} International Conference on Robotics and Automation},
  year         = {2024},
}

@inproceedings{DBLP:conf/siggraph/SahariaCCLHSF022-palette,
  author       = {Chitwan Saharia and
                  William Chan and
                  Huiwen Chang and
                  Chris A. Lee and
                  Jonathan Ho and
                  Tim Salimans and
                  David J. Fleet and
                  Mohammad Norouzi},
  title        = {Palette: Image-to-Image Diffusion Models},
  booktitle    = {ACM Special Interest Group for Computer Graphics and Interactive Techniques},
  year         = {2022},
}

@inproceedings{DBLP:conf/icml/RadfordKHRGASAM21,
  author       = {Alec Radford and
                  Jong Wook Kim and
                  Chris Hallacy and
                  Aditya Ramesh and
                  Gabriel Goh and
                  Sandhini Agarwal and
                  Girish Sastry and
                  Amanda Askell and
                  Pamela Mishkin and
                  Jack Clark and
                  Gretchen Krueger and
                  Ilya Sutskever},
  title        = {Learning Transferable Visual Models From Natural Language Supervision},
  booktitle    = {International Conference on Machine Learning},
  year         = {2021},
}

@inproceedings{DBLP:conf/iccv/ChoiKJGY21-ilvr,
  author       = {Jooyoung Choi and
                  Sungwon Kim and
                  Yonghyun Jeong and
                  Youngjune Gwon and
                  Sungroh Yoon},
  title        = {{ILVR:} Conditioning Method for Denoising Diffusion Probabilistic
                  Models},
  booktitle    = {{IEEE/CVF} International Conference on Computer Vision},
  year         = {2021},
}

@inproceedings{DBLP:conf/iclr/MengHSSWZE22-sdedit,
  author       = {Chenlin Meng and
                  Yutong He and
                  Yang Song and
                  Jiaming Song and
                  Jiajun Wu and
                  Jun{-}Yan Zhu and
                  Stefano Ermon},
  title        = {SDEdit: Guided Image Synthesis and Editing with Stochastic Differential
                  Equations},
  booktitle    = {International Conference on Learning Representations},
  year         = {2022},
}

@inproceedings{DBLP:conf/cvpr/AvrahamiLF22-blended-diffusion,
  author       = {Omri Avrahami and
                  Dani Lischinski and
                  Ohad Fried},
  title        = {Blended Diffusion for Text-driven Editing of Natural Images},
  booktitle    = {{IEEE/CVF} Conference on Computer Vision and Pattern Recognition},
  year         = {2022},
}

@inproceedings{DBLP:conf/cvpr/Chen0SDG18-domain-adaptation-faster-rcnn,
  author       = {Yuhua Chen and
                  Wen Li and
                  Christos Sakaridis and
                  Dengxin Dai and
                  Luc Van Gool},
  title        = {Domain Adaptive Faster {R-CNN} for Object Detection in the Wild},
  booktitle    = {IEEE/CVF Conference on Computer Vision and Pattern Recognition},
  year         = {2018},
}

@incollection{DBLP:series/acvpr/HoffmanTDS17-simultaneous-deep-transfer,
  author       = {Judy Hoffman and
                  Eric Tzeng and
                  Trevor Darrell and
                  Kate Saenko},
  title        = {Simultaneous Deep Transfer Across Domains and Tasks},
  booktitle    = {Domain Adaptation in Computer Vision Applications},
  year         = {2017},
}

@inproceedings{DBLP:conf/aaai/ZhuangHHS20-iFAN,
  author       = {Chenfan Zhuang and
                  Xintong Han and
                  Weilin Huang and
                  Matthew R. Scott},
  title        = {iFAN: Image-Instance Full Alignment Networks for Adaptive Object Detection},
  booktitle    = {{AAAI} Conference on Artificial Intelligence},
  year         = {2020},
}

@inproceedings{DBLP:conf/nips/HoJA20-ddpm,
  author       = {Jonathan Ho and
                  Ajay Jain and
                  Pieter Abbeel},
  title        = {Denoising Diffusion Probabilistic Models},
  booktitle    = {Advances in Neural Information Processing Systems},
  year         = {2020},
}

@inproceedings{DBLP:conf/nips/HeuselRUNH17-fid,
  author       = {Martin Heusel and
                  Hubert Ramsauer and
                  Thomas Unterthiner and
                  Bernhard Nessler and
                  Sepp Hochreiter},
  title        = {GANs Trained by a Two Time-Scale Update Rule Converge to a Local Nash
                  Equilibrium},
  booktitle    = {Advances in Neural Information Processing Systems},
  year         = {2017},
}

@article{DBLP:journals/corr/abs-2505-16360-cactif,
  author       = {Estelle Chigot and
                  Dennis G. Wilson and
                  Meriem Ghrib and
                  Thomas Oberlin},
  title        = {Style Transfer with Diffusion Models for Synthetic-to-Real Domain
                  Adaptation},
  journal      = {CoRR},
  volume       = {abs/2505.16360},
  year         = {2025},
}

@inproceedings{DBLP:conf/nips/WuS0Z24-OSEDiff,
  author       = {Rongyuan Wu and
                  Lingchen Sun and
                  Zhiyuan Ma and
                  Lei Zhang},
  title        = {One-Step Effective Diffusion Network for Real-World Image Super-Resolution},
  booktitle    = {Advances in Neural Information Processing Systems},
  year         = {2024},
}

@misc{Fraundorfer2019DroneDataset-semantic-drone,
  author       = {{Institute of Computer Graphics and Vision, TU Graz, Team Fraundorfer}},
  title        = "{Semantic Drone Dataset – ICG, TU Graz}",
  howpublished = {\url{http://dronedataset.icg.tugraz.at}},
  year         = {2019},
}

@article{akshatha2023manipal,
  title={Manipal-UAV person detection dataset: A step towards benchmarking dataset and algorithms for small object detection},
  author={Akshatha, KR and Karunakar, AK and Satish Shenoy, B and Phani Pavan, K and Chinmay, V Dhareshwar and others},
  journal={ISPRS Journal of Photogrammetry and Remote Sensing},
  year={2023},
}

@article{kuznetsova2020open,
  title={The open images dataset v4: Unified image classification, object detection, and visual relationship detection at scale},
  author={Kuznetsova, Alina and Rom, Hassan and Alldrin, Neil and Uijlings, Jasper and Krasin, Ivan and Pont-Tuset, Jordi and Kamali, Shahab and Popov, Stefan and Malloci, Matteo and Kolesnikov, Alexander and others},
  journal={International journal of computer vision},
  year={2020},
}

@inproceedings{hu2022lora,
title={Lo{RA}: Low-Rank Adaptation of Large Language Models},
author={Edward J Hu and yelong shen and Phillip Wallis and Zeyuan Allen-Zhu and Yuanzhi Li and Shean Wang and Lu Wang and Weizhu Chen},
booktitle={International Conference on Learning Representations},
year={2022},
}

@inproceedings{zhang2024the,
title={The Emergence of Reproducibility and Consistency in Diffusion Models},
author={Huijie Zhang and Jinfan Zhou and Yifu Lu and Minzhe Guo and Peng Wang and Liyue Shen and Qing Qu},
booktitle={International Conference on Machine Learning},
year={2024},
}

@inproceedings{yaras2024compressible,
title={Compressible Dynamics in Deep Overparameterized Low-Rank Learning \& Adaptation},
author={Yaras, Can and Wang, Peng and Balzano, Laura and Qu, Qing},
booktitle={International Conference on Machine Learning},
year={2024},
url={https://openreview.net/forum?id=uDkXoZMzBv}
}

@inproceedings{shen2023progressive,
  title={Progressive transformation learning for leveraging virtual images in training},
  author={Shen, Yi-Ting and Lee, Hyungtae and Kwon, Heesung and Bhattacharyya, Shuvra S},
  booktitle={ IEEE/CVF Conference on Computer Vision and Pattern Recognition},
  year={2023}
}

@article{liu2020uav,
  title={Uav-yolo: Small object detection on unmanned aerial vehicle perspective},
  author={Liu, Mingjie and Wang, Xianhao and Zhou, Anjian and Fu, Xiuyuan and Ma, Yiwei and Piao, Changhao},
  journal={Sensors},
  year={2020},
}

@article{chen2024sim2real,
  title={Sim2Real in reconstructive spectroscopy: Deep learning with augmented device-informed data simulation},
  author={Chen, Jiyi and Li, Pengyu and Wang, Yutong and Ku, Pei-Cheng and Qu, Qing},
  journal={APL Machine Learning},
  year={2024},
}

@inproceedings{yang2024vip,
  title={Vip: Versatile image outpainting empowered by multimodal large language model},
  author={Yang, Jinze and Wang, Haoran and Zhu, Zining and Liu, Chenglong and Wu, Meng and Sun, Mingming},
  booktitle={Asian Conference on Computer Vision},
  year={2024}
}

@inproceedings{lugmayr2022repaint,
  title={Repaint: Inpainting using denoising diffusion probabilistic models},
  author={Lugmayr, Andreas and Danelljan, Martin and Romero, Andres and Yu, Fisher and Timofte, Radu and Van Gool, Luc},
  booktitle={IEEE/CVF Conference on Computer Vision and Pattern Recognition},
  year={2022}
}

@article{chen2024exploring,
  title={Exploring low-dimensional subspace in diffusion models for controllable image editing},
  author={Chen, Siyi and Zhang, Huijie and Guo, Minzhe and Lu, Yifu and Wang, Peng and Qu, Qing},
  journal={Advances in neural information processing systems},
  year={2024}
}

@article{song2020denoising,
  title={Denoising diffusion implicit models},
  author={Song, Jiaming and Meng, Chenlin and Ermon, Stefano},
  journal={arXiv preprint arXiv:2010.02502},
  year={2020}
}

@article{ye2023ip,
  title={Ip-adapter: Text compatible image prompt adapter for text-to-image diffusion models},
  author={Ye, Hu and Zhang, Jun and Liu, Sibo and Han, Xiao and Yang, Wei},
  journal={arXiv preprint arXiv:2308.06721},
  year={2023}
}

@book{sanders2016introduction,
  title={An introduction to Unreal engine 4},
  author={Sanders, Andrew},
  year={2016},
  publisher={AK Peters/CRC Press}
}

@article{goodfellow2020generative,
  title={Generative adversarial networks},
  author={Goodfellow, Ian and Pouget-Abadie, Jean and Mirza, Mehdi and Xu, Bing and Warde-Farley, David and Ozair, Sherjil and Courville, Aaron and Bengio, Yoshua},
  journal={Communications of the ACM},
  year={2020},
}

@inproceedings{rombach2022high,
  title={High-resolution image synthesis with latent diffusion models},
  author={Rombach, Robin and Blattmann, Andreas and Lorenz, Dominik and Esser, Patrick and Ommer, Bj{\"o}rn},
  booktitle={IEEE/CVF conference on computer vision and pattern recognition},
  year={2022}
}

@article{chen2021understanding,
  title={Understanding domain randomization for sim-to-real transfer},
  author={Chen, Xiaoyu and Hu, Jiachen and Jin, Chi and Li, Lihong and Wang, Liwei},
  journal={arXiv preprint arXiv:2110.03239},
  year={2021}
}

@inproceedings{zhang2019multimodal,
  title={Multimodal style transfer via graph cuts},
  author={Zhang, Yulun and Fang, Chen and Wang, Yilin and Wang, Zhaowen and Lin, Zhe and Fu, Yun and Yang, Jimei},
  booktitle={IEEE/CVF International Conference on Computer Vision},
  year={2019}
}

@inproceedings{wu2024seesr,
  title={Seesr: Towards semantics-aware real-world image super-resolution},
  author={Wu, Rongyuan and Yang, Tao and Sun, Lingchen and Zhang, Zhengqiang and Li, Shuai and Zhang, Lei},
  booktitle={IEEE/CVF conference on computer vision and pattern recognition},
  year={2024}
}

\clearpage
\appendix
\addcontentsline{toc}{section}{Appendix}


\section{More Visualized Results}
\label{sec:morequalitativeresult}

\begin{figure}[h]
    \centering 
    \includegraphics[width=0.57\textwidth, height=0.74\textheight]{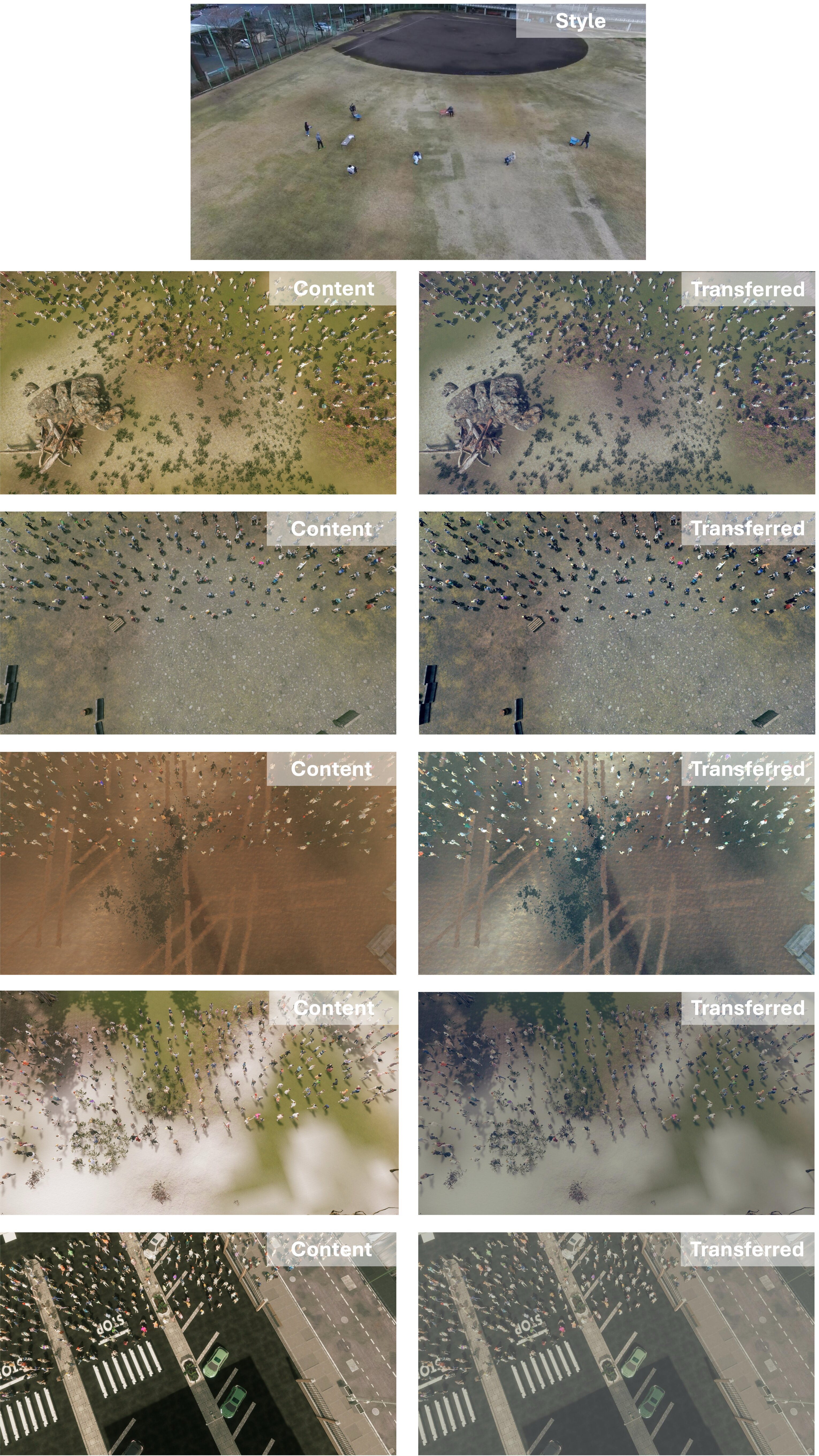} 
    \caption{More visualized results of our adapted images} 
    \label{fig:morevis} 
\end{figure}

\clearpage
\section{Comparison with Adaptive Detector Methods}
\label{sec:morequantitativeresults}

\begin{table}[h]
\centering
\resizebox{\textwidth}{!}{%
\begin{tabular}{l|cc|cc|cc|cc}
\toprule
\multirow{2}{*}{\textbf{Adapted Detectors}} & \multicolumn{2}{c|}{\textbf{Okutama}} & \multicolumn{2}{c|}{\textbf{Semantic-Drone}} & \multicolumn{2}{c|}{\textbf{VisDrone}} & \multicolumn{2}{c}{\textbf{Manipal-UAV}} \\
\cmidrule(lr){2-3} \cmidrule(lr){4-5} \cmidrule(lr){6-7} \cmidrule(lr){8-9} 
  & \textbf{mAP50} & \textbf{mAP95} & \textbf{mAP50} & \textbf{mAP95} & \textbf{mAP50} & \textbf{mAP95} & \textbf{mAP50} & \textbf{mAP95} \\
\midrule


DA Faster R-CNN  & 48.26 & 17.43 & 24.53 & 9.81 & 19.75 & 7.92 & 42.38 & 17.61 \\
SWDA          & 53.92 & 18.17 & 27.39 & 12.73 & 25.63 & 13.89 & 51.44 & 21.34 \\
DA-DETR         & 55.19 & 18.92 & 31.21 & 14.13 & 26.87 & 14.11 & 52.20 & \textbf{21.96} \\
\textbf{Ours}  & \textbf{68.20} & \textbf{25.30} & \textbf{32.70} & \textbf{15.00} & \textbf{31.50} & \textbf{15.10} & \textbf{53.80} & 20.50 \\

\bottomrule
\end{tabular}
}
\vspace{0.3cm}
\caption{Performance comparison with domain adaptive detection methods.}
\label{tab:adaptive_results}
\end{table}

\section{Stage 2 Training}
\label{sec:stage2training}
\paragraph{Training of the Local Refinement Module.}
To restore high-frequency details after global style transfer, we train the Local Refinement (LR) module as a one-step diffusion-based super-resolution network following the design of OSEDiff~~\autocite{DBLP:conf/nips/WuS0Z24-OSEDiff}. 
Specifically, we construct training pairs by extracting object patches from real UAV images using ground-truth bounding boxes as high-quality targets $x_H$, while their degraded counterparts $x_L$ are generated via bicubic downsampling and mild blur/noise perturbations to mimic the GST outputs.

The LR module adopts a Stable Diffusion backbone with LoRA adaptation. 
We insert trainable LoRA layers into the VAE encoder $\mathcal{E}$ and diffusion UNet $\bm{\epsilon}_\theta$, producing $\mathcal{E}'$ and $\bm{\epsilon}_\theta$, while keeping the VAE decoder $\mathcal{D}$ fixed to preserve the latent space consistency. 
Given a degraded patch $x_L$, we first encode it as $z_L=\mathcal{E}'(x_L)$ and perform a single reverse diffusion step at timestep $T$:
\begin{align}
z_H = \frac{z_L - \beta_T\,\bm{\epsilon}_\theta(z_L;T,\bm{c}_y)}{\alpha_T}, 
\qquad 
\hat{x} = \mathcal{D}(z_H),
\end{align}
where $\bm{c}_y$ is a tag-style prompt extracted from the input via DAPE~~\autocite{wu2024seesr}. 
Unlike conventional diffusion SR methods that start from random noise and require multiple sampling steps, this formulation directly uses the degraded image latent as the starting point and performs only one denoising step.

The training objective follows OSEDiff and combines a data fidelity term and a variational score distillation (VSD) regularization:
\begin{equation}
L = L_{\text{MSE}}(\hat{x},x_H) + \lambda_1 L_{\text{LPIPS}}(\hat{x},x_H) + \lambda_2 L_{\text{VSD}} .
\end{equation}
The VSD loss aligns the predicted noise of the trainable diffusion network with that of a frozen teacher model, ensuring that the refined outputs follow the natural image distribution while enabling efficient one-step inference. 
During both training and inference, the latent update is restricted to the predefined instance mask so that refinement only affects object regions.

\section{FID Comparison}
\label{sec:FID}

To further explore weather our method makes the image-wise and patch-wise distribution closer to target data, we use Fréchet inception distance (FID) as a metric to evaluate the distance between transferred data distribution and target data distribution. It is worth noting that our method mainly decrease the FIDs of the patch-wise instances, and it inspires us to further explore what is the conclusive factor when training a detector.

\begin{table}[h]
\centering
\resizebox{\linewidth}{!}{
\begin{tabular}{l|cccc|cccc}
\toprule
\textbf{Method} 
& \multicolumn{4}{c|}{\textbf{Patch-wise FID} $\downarrow$} 
& \multicolumn{4}{c}{\textbf{Image-wise FID} $\downarrow$} \\
& Okutama & Semantic & VisDrone & Manipal & Okutama & Semantic & VisDrone & Manipal\\
\midrule
Synthetic(Baseline)     & 104 & 120 & 142 & 147 & 320 & 355 & 273 & 246\\
GST only                & 107 & 118 & 139 & 144 & 235 & 290 & 256 & 234\\
LR only                 & 76  & 92  & 111 & 125 & 336 & 387 & 268 & 241\\
HR only                 & 97  & 112 & 134 & 154 & 377 & 423 & 281 & 264\\
GST + LR                & 69  & 95  & 102 & 113 & 323 & 340 & 244 & 229\\
GST + LR + HR           & 62  & 87  & 99  & 117 & 267 & 288 & 257 & 196\\
\bottomrule
\end{tabular}
}
\vspace{0.3cm}
\caption{FID comparison of different components in the CFHA framework on datasets. Lower is means closer to the target distribution.}
\label{tab:fid_ablation}
\end{table}

\section{Additional Related Work}
\label{RelatedWork}
\subsection{Sim2Real Adaptation for UAV-based human Detection}

To mitigate data scarcity, many works leverage either advanced algorithms ~\autocite{wang2024convolution, liu2020uav} or large-scale synthetic datasets~\autocite{shen2023progressive, chen2024sim2real} with free annotations to enhance UAV perception tasks, including object detection and semantic segmentation. 
However, the significant appearance gap between synthetic and real UAV imagery makes effective simulation-to-real (Sim2Real) domain adaptation essential. Additionally, synthetic data may introduce performance degradation~\autocite{shi2025closer}, requiring more careful treatment and utilization in model training.

To bridge the Sim2Real gap, several works explore image-level or feature-level adaptation techniques. For instance, ~\autocite{DBLP:journals/ral/BarisicPB22-sim2air} introduces Sim2Air, a domain adaptation framework for cross-weather UAV tracking, which incorporates spatiotemporal attention to mitigate style and temporal drift. Their work highlights the challenge of temporal consistency in real-world UAV deployment. ~\autocite{DBLP:journals/ral/TruongCB21-bi-directional} proposed a bi-directional simulation-to-reality approach for object detection, leveraging dual generative models to jointly enhance synthetic realism and improve detector adaptation. This method emphasizes the hierarchical feedback between detection quality and image fidelity. ~\autocite{DBLP:conf/icra/SaadiyeanSS24-multi-scale} develops a multi-scale adversarial adaptation framework for UAV pedestrian detection, using feature-level alignment across scales to improve robustness under varying UAV altitudes and viewpoints, underscoring the importance of spatial resolution modeling in aerial detection.

Despite the growing interest in Sim2Real adaptation, relatively few results have specifically addressed the Sim2Real gap for UAV-based human object detection. One notable exception is PTL~~\autocite{shen2023progressive}, which progressively transforms the local humans in synthetic images into realistic ones using a conditional GAN framework. In contrast, our framework transforms the images from both global and local perspectives and utilizes diffusion models as the generative method.

\subsection{Diffusion Models for Image-to-Image Translation}

Diffusion models ~\autocite{zhang2024the,DBLP:conf/nips/HoJA20-ddpm,DBLP:conf/iclr/PodellELBDMPR24,DBLP:conf/cvpr/RombachBLEO22} have recently emerged as a powerful class of generative models, demonstrating superior performance in diverse image synthesis tasks, including unconditional ~\autocite{DBLP:conf/nips/HoJA20-ddpm, song2020denoising}, inpainting~\autocite{yang2024vip, lugmayr2022repaint}, and conditional translation ~\autocite{DBLP:conf/iclr/MengHSSWZE22-sdedit, chen2024exploring}. Their iterative denoising formulation avoids common issues in GAN-based models such as mode collapse and training instability, making them particularly appealing for high-fidelity image translation. Chitwan et al.~~\autocite{DBLP:conf/siggraph/SahariaCCLHSF022-palette} proposed Palette, a conditional diffusion model tailored for image-to-image tasks such as colorization and inpainting. Jooyoung et al.~\autocite{DBLP:conf/iccv/ChoiKJGY21-ilvr} introduced ILVR, which guides the diffusion process using low-resolution conditioning, allowing for semantically aligned outputs while preserving local structures. Their work showed that diffusion models can effectively preserve the content structure when guided by weak conditions. Chenlin et al.~\autocite{DBLP:conf/iclr/MengHSSWZE22-sdedit} developed Contextual Diffusion Models (CDM) for controllable image synthesis by injecting semantic maps or sketches as spatial conditions. 
Omri et al.~\autocite{DBLP:conf/cvpr/AvrahamiLF22-blended-diffusion} proposed Blended Diffusion, a technique for localized image editing via spatially modulated noise, enabling region-specific transformations while keeping the rest of the image unchanged. This aligns with our objective of performing localized refinement in Sim2Real transfer while retaining structural consistency.

\vspace{2pt}
Despite these advances, existing diffusion-based translation methods primarily focus on artistic editing or general image synthesis tasks, and do not explicitly address domain adaptation or the hierarchical decomposition of distribution shifts. In contrast, our proposed HSRA framework leverages two dedicated diffusion modules for global style alignment and local structure refinement, enabling effective Sim2Real transfer in detection-centric tasks under a unified and interpretable architecture.

\vspace{-5pt}
\subsection{Hierarchical Disentanglement for Domain Transfer}

Visual domain adaptation methods often treat the domain gap between source and target domains as a single transformation, applying global alignment techniques such as adversarial training or feature matching. However, recent research has recognized that domain shift may occur at multiple semantic levels—e.g., in style (color, illumination) and structure (object geometry, fine textures)—and that explicitly modeling such hierarchy can lead to more robust and interpretable adaptation. Yuhua et al.~~\autocite{DBLP:conf/cvpr/Chen0SDG18-domain-adaptation-faster-rcnn} proposed a hierarchical feature alignment (CFHA) method, aligning features at both global and class-specific levels across domain layers. By modeling feature shift hierarchically, HFA improves generalization in semantic segmentation and object detection tasks. Judy et al.~~\autocite{DBLP:series/acvpr/HoffmanTDS17-simultaneous-deep-transfer} pioneered class-wise feature alignment for visual recognition, proposing that inter-class and intra-class discrepancies should be addressed separately. Their framework serves as an early instance of disentangled domain alignment. Chenfan et al.~~\autocite{DBLP:conf/aaai/ZhuangHHS20-iFAN} proposed iFAN, a hierarchical alignment framework for unsupervised domain adaptive object detection that integrates both image-level and instance-level alignment. By designing category-aware and category-correlation instance alignment modules, iFAN effectively models multi-level domain shifts and achieves remarkable results on domain adaptation datasets: SIM10K to Cityscapes and Cityscapes to Foggy Cityscapes.
\vspace{2pt}
While these methods explore hierarchical alignment at different hierarchical structures, they do not incorporate image-level generative modeling to handle domain shifts in pixel space. Our proposed CFHA framework advances this line of work by employing two diffusion-based model to collaborately transfer global style and refine local details, which is more suitable for UAV-based images, resulting in a more fine-grained and controllable Sim2Real adaptation pipeline for UAV detection.


\end{document}